\pdfoutput=1

\documentclass[11pt]{article}

\usepackage[]{ACL2023}

\usepackage{times}
\usepackage{latexsym}

\usepackage[T1]{fontenc}

\usepackage[utf8]{inputenc}

\usepackage{microtype}

\usepackage{inconsolata}

\usepackage{color} %
\usepackage{comment} %
\usepackage{graphicx} %
\usepackage{booktabs} %
\usepackage{multirow} %
\usepackage{amssymb} %
\usepackage{soul}
\usepackage[linesnumbered,ruled,vlined]{algorithm2e}
\usepackage{amsmath}
\usepackage[export]{adjustbox}
\usepackage{enumitem} %
\usepackage{colortbl} %
\usepackage{xcolor} %
\usepackage[toc,page,header]{appendix}
\usepackage{minitoc}

\graphicspath{images}

\DeclareMathOperator*{\argmax}{arg\,max}

\usepackage{xspace} %

\newcommand{\ood}{\textsc{ood}\xspace}
\newcommand{\ind}{\textsc{ind}\xspace}

\newcommand{\ho}{\ensuremath{h_{0}}}
\newcommand{\viscontext}{\ensuremath{\textbf{v}}}

\newcommand{\siter}{\ensuremath{st_{adp}}}
\newcommand{\lradapt}{\ensuremath{lr_{adp}}}

\newcommand{\truetarg}{\ensuremath{t_g}}

\newcommand{\simpred}{\ensuremath{t_{sim}}}

\newcommand{\simout}{\ensuremath{o_{sim}}}

\newcommand{\utt}{\ensuremath{u}}

\title{Speaking the Language of Your Listener:\\Audience-Aware Adaptation via Plug-and-Play Theory of Mind}

\newcommand{\amsterdam}[0]{$^{\triangleleft}$}
\newcommand{\rome}[0]{$^{\diamond}$}

\author{Ece Takmaz\amsterdam \thanks{\ \ Shared first authorship.}\ , Nicolo' Brandizzi\rome \footnotemark[1]\ , Mario Giulianelli\amsterdam, Sandro Pezzelle\amsterdam, Raquel Fernández\amsterdam\\
 \amsterdam University of Amsterdam \rome Sapienza University of Rome\\
	\texttt{\{ece.takmaz|m.giulianelli|s.pezzelle|raquel.fernandez\}@uva.nl}\\
	\texttt{brandizzi@diag.uniroma1.it}}
\begin{document}
\maketitle

\begin{abstract}

Dialogue participants may have varying levels of knowledge about the topic under discussion. In such cases, it is essential for speakers to adapt their utterances by taking their audience into account. Yet, it is an open question how such adaptation can be modelled in computational agents. In this paper, we model a visually grounded referential game between a knowledgeable speaker and a listener with more limited visual and linguistic experience. Inspired by psycholinguistic theories, we endow our speaker with the ability to adapt its referring expressions via a simulation module that monitors the effectiveness of planned utterances from the listener's perspective. We propose an adaptation mechanism building on plug-and-play approaches to controlled language generation, where utterance generation is steered on the fly by the simulator without finetuning the speaker's underlying language model. Our results and analyses show that our approach is effective: the speaker's utterances become closer to the listener's domain of expertise, which leads to higher communicative success. 

\end{abstract}

\section{Introduction}
\label{sec:intro}

Speakers tend to adapt their language use to the perceived knowledge, information, and linguistic abilities of their interlocutors \cite{isaacs1987references,Clark1996,Pickering2004-PICTAM}. When adults speak with children, for example, they use simplified expressions to ensure children are able to understand~\cite{saxton2009inevitability}; when computational linguists give a talk at a cognitive science conference, they (hopefully) avoid making extensive use of NLP jargon, as that would prevent their audience from following through the presentation. 
Successful adaptation to the conceptual knowledge of conversational partners requires the ability to represent and reason about others’ mental states \cite{tomasello2005constructing}, a socio-cognitive ability typically referred to as Theory of Mind \cite[ToM;][]{premack1978tom}. %
Yet, speakers do not always resort to explicitly modelling the knowledge of their dialogue partner: due to different cognitive costs and pressures, they sometimes plan their utterances egocentrically, i.e., only taking into account their own knowledge and abilities \cite{keysar2007communication}. 

\begin{figure}[t]
	\centering
	\includegraphics[width=.9\columnwidth]{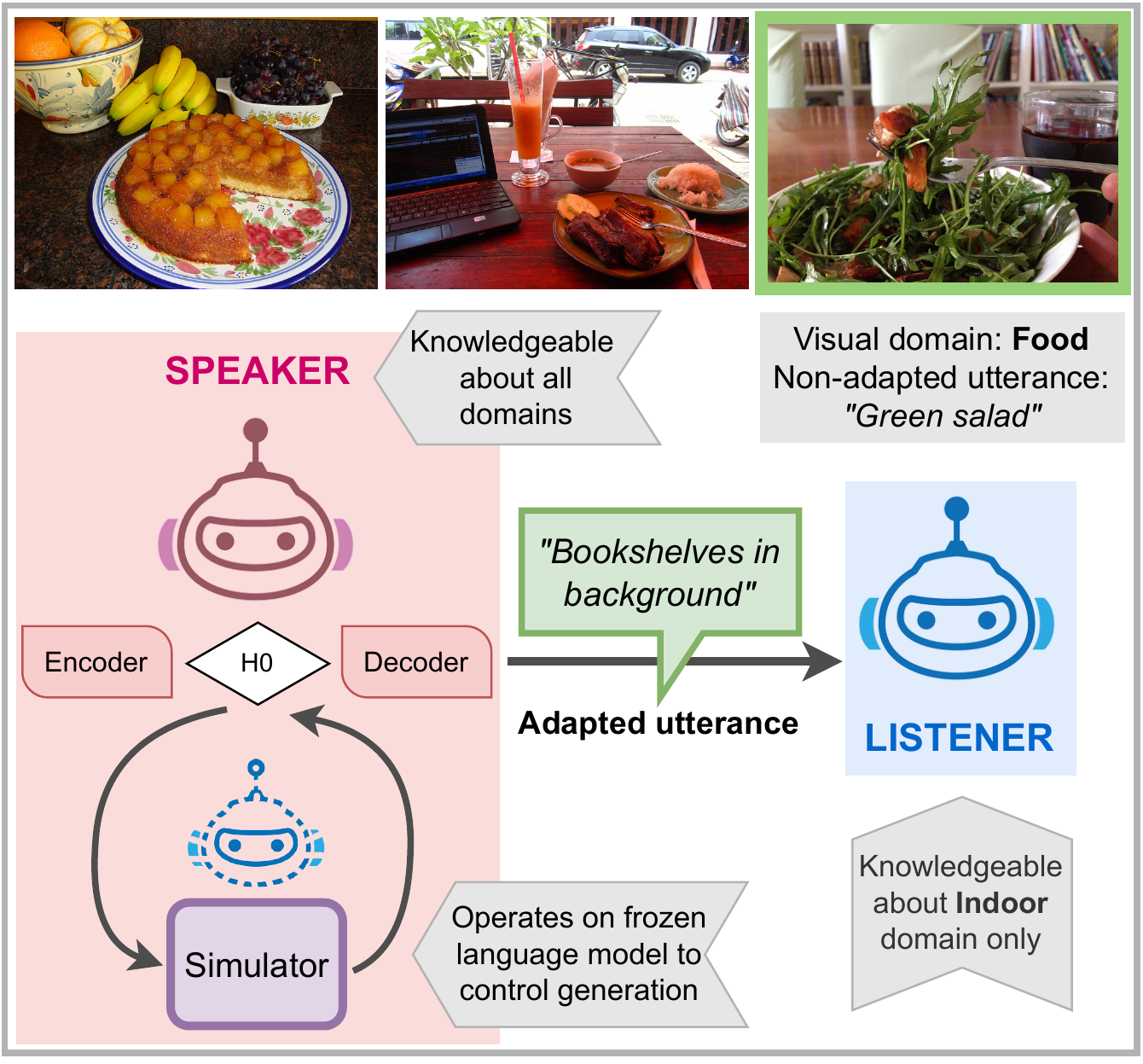} %
	\caption{
    An illustration of our knowledge-asymmetric setup where an expert \textit{Speaker} interacts with a less knowledgeable \textit{Listener}. 
    The \textit{Speaker} tailors its utterance about an image from the \textbf{food} domain for a \textit{Listener} who only knows about the \textbf{indoor} domain. %
    The speaker's \textit{Simulator} module inspired by Theory of Mind guides this adaptation. The adapted utterance exploits indoor terms (\textit{`bookshelves}') without referring to food.}
    \label{fig:pipeline}
\end{figure}

In this paper, we model a communicative situation where the interlocutors have \textit{asymmetric language abilities}: a proficient speaker interacts with a listener characterised by limited semantic knowledge to complete a reference game, as illustrated in Fig.~\ref{fig:pipeline}.
Our goal is to mimic a scenario in which, for example, a high school physics professor can make complex atomic models understandable to young students by using terminology familiar to them, such as culinary terminology to explain Thomson's `plum pudding model'. %
We focus on the speaker's Referring Expression Generation~\citep[REG;][]{reiter1997building,krahmer-van-deemter-2012-computational} in a multimodal dialogue setting and use REG models equipped with visual perception %
to generate discriminative image descriptions within a set of related image candidates. 
Several psycholinguistic theories have proposed that language production is interwoven with comprehension via `forward prediction'---i.e., producing an utterance involves predicting how a comprehender would understand it 
\cite[e.g.,][]{pickering2013integrated,roelofs2020self}. 
Inspired by this idea, we equip our speaker model with a \textit{simulator}, i.e., a %
module that `simulates' whether a listener would be able to identify the target referent. Based on this predicted behaviour (i.e., the expected effect of the planned utterance), the simulator modifies the generation plan on the fly to increase communicative success.

These are the main contributions of our study:\footnote{Code and models available at \url{https://github.com/nicofirst1/speaker-adaptation}}

\begin{itemize}[leftmargin=11pt,itemsep=-0.5pt]

\item We model adaptation between agents with asymmetric knowledge, using a referential task as case study, where agents communicate in natural language about realistic images (in contrast to related work using synthetic data---see \textsection\ref{sec:related-work}).  

\item  We propose a novel simulation-based approach and test it in two settings: 
(1) a \textit{self-aware} setting where the speaker predicts how a generic listener (with the same knowledge as the speaker) would resolve a planned utterance, and  
(2) an \textit{audience-aware} 
setting where %
the speaker learns---from the behaviour of a listener with restricted semantic knowledge---to form representations of the listeners' knowledge %
\cite{clark1985language,isaacs1987references} and predict their responses.\looseness-1

\item We exploit the simulator's representations in an innovative way: by leveraging a \textit{plug-and-play} approach %
originally introduced for controllable text generation~\cite{dathathri2020plug}, which steers language production at the decoding stage without altering the underlying language model.

\item We show that our approach leads to increased resolution accuracy; in particular, our audience-aware speaker is able to adapt its utterances effectively when referring to a target within a visual domain unknown to the listener. 

\item We provide an in-depth analysis of the patterns present in the adapted utterances and the model's production strategies underpinning our results. 

\end{itemize}

\section{Related Work}
\label{sec:related-work}

\subsection{Pragmatic Reference Generation}
\label{sec:related-work-pragmatic}

Speakers tend to design their referring expressions to be pragmatically informative, i.e., discriminative from the listener's perspective. Most approaches to pragmatic reference expression generation (REG) have considered scenarios where we can assume a shared set of linguistic conventions between speakers and addressees (common domain and training data). The Rational Speech Act framework \cite[RSA;][]{frank2012predicting,goodman2013knowledge,goodman2016pragmatic} has become a popular option for characterising such settings, with REG models that reason probabilistically about their interlocutors' interpretation via recursively defined speaker and listener models \cite{andreas-klein-2016-reasoning,monroe-etal-2017-colors,cohn-gordon-etal-2018-pragmatically,zarriess-schlangen-2019-know,fried-etal-2021-reference}, possibly taking into account information accumulated during interaction \cite{hawkins2020continual}. 
There also exist joint speaker-listener models that are not recursive in the RSA sense. In these models, speakers can become listener-aware at inference time thanks to enhanced decoding algorithms \cite{vedantam2017context} or they can learn to generate discriminative utterances at training time, for example via altered supervised training objectives
\cite{mao2016generation} or auxiliary reinforcement learning (RL) modules \cite{Yu2017AJS}, including approaches where the RL rewards are determined by the reference resolution success of a listener model \cite{lazaridou-etal-2020-multi}. 

Our model, too, produces audience-aware discriminative image descriptions through an auxiliary module that captures the listener's perspective. However, 
in contrast to the above studies, the setting we investigate has two distinct key features: 
(1) we model situations with \textit{knowledge asymmetry} between the dialogue participants,
and (2) we experiment with \textit{plug-and-play controlled generation} methods that result in temporary updates to the speaker's language model---rather than steering generation via %
recursive probabilistic reasoning.
We review work related to these two aspects next.

\subsection{Knowledge Asymmetry \& Referring Tasks}
\label{sec:related-work-asymmetry}

What if the speaker and the listener have access to differing semantic knowledge? It is well known that speakers are able to adapt to less proficient addressees \cite{isaacs1987references}. 
\citet{janarthanam2010learning} were one of the first to address adaptation in dialogue systems with asymmetric knowledge. They modelled REG for technical domains where users may not know the jargon, using RL to learn a REG policy from a user simulation.  %
More recently, \citet{ohashi-higashinaka-2022-adaptive} focus on generating utterances in task-oriented dialogue with users that have limited vocabulary. They exploit the natural language understanding module of the system (representing user understanding) to set up a reward function, which is then used to finetune the NLG module via RL. 

In the context of visually grounded referring tasks, 
\citet{bao-etal-2022-learning} focus on a scenario where the listener has comprehension difficulties and model adaptation by reweighing the probability of candidate referring utterances as a function of their likelihood to be successfully interpreted by the listener. Similarly, \citet{liu-etal-2016-evaluate} apply ToM-based listener modelling, where the speaker generates multiple candidate utterances and ranks them with the help of the ToM listener.  Generating and ranking multiple utterances, however, is an inefficient %
production mechanism. For these reasons, others have tried to condition the speaker model prior to utterance generation, mainly with external modules. \citet{corona2019modeling} model interactions where the listener has an impaired perceptual system and implement this conditioning through an external policy network that takes as input listener embeddings. While \citet{zhu2021few} propose a ToM module that tracks the listener's understanding via meta-learning for few-shot coordination in a setup where listeners understand different languages. \citet{singh-etal-2023-know} train an attention-based adapter layer in a reward-based manner as part of a multi-agent referential game where the speaker aims to generate utterances that would be understood by one listener, but not the other. Finally, \citet{sheadapts} have a setup that is the most similar to ours, where Expert speakers adapt to Layman listeners. But unlike our \textit{plug-and-play} approach, the authors follow the RSA framework in developing audience-aware models that are updated through interaction.

\subsection{Adaptive Controlled Generation}
\label{sec:related-work-control}

Most of the approaches to adaptation we have reviewed apply RL to the speaker model or  finetune its language model through interaction. As a result, the speaker is not able to retain its original knowledge, which might cause catastrophic forgetting~\citep{MCCLOSKEY1989109,FRENCH1999128}. %
With the advent of large pretrained language models, a plethora of new methods for controlled text generation have been proposed, including prefix-tuning \citep{prefix, pada}, prompting \citep{gpt3}, adapters \citep{houlsby19a, adapterhub, madx}, and energy-based constraints \cite{qin2022cold}. Visual prefixes and prompts \citep{flamingo} have also been used to condition generation, especially without training the full language model.  

We argue that this recent line of research offers promising alternative frameworks for adaptive REG. In particular, we investigate a solution to adaptation inspired by the \textit{plug-and-play} approach to controlled text generation \cite[PPLM;][]{dathathri2020plug,pascual-etal-2021-plug-play},
which has been used to steer large pretrained language models towards generating texts with certain features (e.g., positive/negative sentiment or a given vocabulary distribution). In \citet{dathathri2020plug}, latent representations are updated at inference time with the help of a classifier while keeping the model parameters unchanged. Building on this idea, we propose a modular approach to REG adaptation in asymmetric knowledge settings where a module trained to predict the listener's behaviour---similar to the `prediction net' in the machine ToM model by \citet{abs-1802-07740}---is exploited to control generation on the fly.

\section{Problem Formulation}
\label{sec:problem}

We provide an abstract overview of the problem we address and our approach. Details on the data and the experimental pipeline are given in \textsection\ref{sec:data} and \textsection\ref{sec:models}.

\paragraph{Scenario}
Our setup is a classic referential game: two artificial agents, a speaker and a listener, share a visual context involving multiple images. The speaker produces an utterance to refer to one of the images (the target) and the listener attempts to identify the referent given that utterance. In particular, we model a scenario with \textit{knowledge asymmetry}, where the speaker is more knowledgeable than the listener. We hypothesise that, in such a setup, for communication to be successful, the speaker will need to adapt its utterances to the listener's representational space and language. To make this possible, we endow the speaker with a simulation module and an adaptation mechanism. 

\paragraph{Simulation}
We provide the speaker with a module that simulates how a listener would process a planned utterance. We assume that, by having interacted with listeners in the past, the speaker has learned a model of certain listener types (e.g., a prototypical idea of what a 3-year-old would understand). We operationalise this by pretraining several instances of the simulator, one per listener type, to predict how a listener is likely to resolve a referring utterance. 
We compare three settings:
\begin{itemize}[leftmargin=10pt,itemsep=1pt]

\item \textit{Baseline}: No simulation takes place.

\item \textit{Self-aware}: The simulator is trained to predict how a listener with the same knowledge as the speaker would resolve an utterance. This is equivalent to a pragmatic speaker who reasons about the effect of its utterances on a generic listener (see \textsection\ref{sec:related-work-pragmatic}), but in our approach at test time the listener's interpretations are predicted rather than directly observed.
Our proposal is also inspired by human production models based on `self-monitoring' \cite{levelt1993speaking,roelofs2020self}. 

\item \textit{Audience-aware}: The simulator is trained to predict how a listener with a subset of the speaker's knowledge, i.e., a single domain, would resolve an utterance. Thus, the speaker learns a model---a theory of mind---of a less knowledgeable listener type that allows the speaker to make predictions about the listener's behaviour. When performing the referential task, we assume that the speaker knows the type of the listener beforehand, i.e., which simulator needs to be engaged (similarly to knowing that we are addressing a 3-year-old, for example). 
 
\end{itemize}

\paragraph{Adaptation}
Rather than finetuning the speaker's language model, we exploit the pretrained simulators to control utterance generation on the fly via a monitoring loop. The simulator checks whether planned utterances would be effective; if that is not the case, a loss is backpropagated to update the initial hidden state of the speaker's decoder and a new utterance is generated. Our hypothesis is that such a mechanism will lead to referring utterances that are adapted to the listener's knowledge.

\section{Data}
\label{sec:data}

As a basis for our experiments, we use the PhotoBook dataset~\citep[PB;][]{haber2019photobook}, a collection of task-oriented visually grounded English dialogues between pairs of participants who communicate via written chat. In a PhotoBook game, two participants see their own private sets (`photobooks') of real-life images belonging to the same visual domain. The goal of the interaction is for them to find out which images they have in common. This elicits referring utterances such as  \textit{``I have a little boy holding a phone to a teddy bear''}, where participants refer to an image that their dialogue partner needs to identify among six similar images. Our focus is on the generation of such referring utterances, leaving aside the dialogue context for simplicity. We use the dataset of referring utterances automatically extracted from the PB dialogues by \citet{takmaz-etal-2020-refer}, which includes 41,340 utterances paired with their target image and the other five images in the visual context. 

We choose PhotoBook because its visual contexts consist of realistic images and feature multiple challenging distractors, all selected from the visual domain of the target image. The original images are taken from the Microsoft COCO dataset~\citep{lin2014coco} and belong to 30 different visual domains (e.g., \textit{`person-umbrella'}, \textit{`car-motorcycle'}). To model speaker adaptation to different semantic domains, we split the dataset of PB referring utterances according to the visual domain of each game. We cluster the image domains as a function of the similarity between their vocabulary vectors, constructed by counting word frequencies in the referring utterances belonging to a given domain. We obtain a set of 5 macro-domains (\textit{appliances}, \textit{food}, \textit{indoor}, \textit{outdoor}, \textit{vehicles}), selected so that the domain vocabularies have minimal overlap.
For each cluster of visual domains, we extract the corresponding referring utterance and visual context. We then randomly split these into training (70\%), validation (15\%), and test set (15\%). We also merge the 5 domain-specific datasets into an \textit{`all-domains'} dataset to be used to train domain-general models as described in \textsection\ref{sec:models}. See summary in Table~\ref{tab:data-stats}.
\begin{table}[t]
    \centering
    \resizebox{\columnwidth}{!}{
    \begin{tabular}{@{}c|cccccc@{}}
    \toprule
        \textbf{Domain} & \textbf{Prop} & \textbf{$N$} & \textbf{$|V|$}  & \textbf{Images} & \textbf{Specific} & \textbf{Overlap}\\ \midrule
        \textit{Appliances} & \ 9.4\%  & \ 4,310 & 1,271  & \ 36 & 29.5\% & 23.2\% (\textit{Ind})\\
        \textit{Food}       & 12.4\%   & \ 5,682 &  1,646 & \ 36 & 43.3\% & 22.9\% (\textit{App}) \\
        \textit{Indoor}     & 26.4\%   & 12,088  &  2,477 & \ 96 & 44.3\% & 26.0\% (\textit{Out})\\
        \textit{Outdoor}    & 35.9\%   & 16,427  &  2,858 & 108  & 47.0\% & 26.2\% (\textit{Veh})\\
        \textit{Vehicles}   & 15.8\%   & \ 7,234 &  1,738 & \ 48 & 36.0\% & 26.2\% (\textit{Out}) \\
        \textit{All}        & \ 100\%  & 45,741  &  6,038 & 324  & - & - \\\bottomrule
    \end{tabular}
    }
    \caption{Statistics of the domain-specific datasets: \# of utterances ($N$) and \% within the entire dataset (Prop), vocabulary size ($|V|$), \# of unique images (Images), \% of domain-specific vocabulary (Specific), and max.\ lexical overlap with another domain (Overlap). The max.\ overlap is  between \textit{outdoor} and \textit{vehicles}. Example shared words are \textit{`left'}, \textit{`black'}, \textit{`driving'}, and \textit{`glasses'}.}
    \label{tab:data-stats}
\end{table}

\section{Experimental Pipeline} %
\label{sec:models}

As described in \textsection\ref{sec:problem}, our experimental pipeline includes two agents---a speaker and a listener---implemented as a generative language model instantiating the speaker, a discriminative model instantiating the listener, and a third model, a simulator used by the speaker to assess the forward effect of its planned utterance on the listener. The language model and the discriminator model are adapted from those by \citet{takmaz-etal-2020-refer}, and the simulator model is built on the discriminator's architecture with additional components. We train these models from scratch to have full control over the linguistic and visual knowledge of the agents and their degree of asymmetry. We use ResNet-152 to encode the images~\cite{resnet2016}. See Appendix~\ref{sec:app:train_detail} for more information about the training schemes and hyperparameters.

\subsection{Generative Language Model}
\label{sec:speaker}

The speaker is a visually conditioned language model that generates an utterance describing a target image within a visual context. The model follows an encoder-decoder architecture consisting of a visual encoder that represents the visual context along with the target image, and a decoder for language generation. The decoder generates a referring utterance via nucleus sampling~\citep{Holtzman2020The}, also paying attention to the encoder output at every time step. See Appendix~\ref{sec:app:train_detail:speak} for more details about the model architecture.

We train the visually conditioned language model from scratch using the training set including all domains in PB and optimize the model with respect to Cross Entropy Loss using Adam~\citep{Adam}. We select the best model based on its performance on a set of natural language generation metrics on the validation set. The weights of the trained speaker are then frozen and used as the core language generation model in all our experiments identically.

\paragraph{Performance} 
The speaker's language model obtains reasonable scores 
in terms of classic natural language generation metrics:\footnote{Comparable to those obtained by \citet{takmaz-etal-2020-refer} with their `Ref' model.} 23.8 BLEU-2, 32.9 ROUGE, 44.1 CIDEr, and 57.7 BERTScore F1~\citep{Papineni:2002, Lin2004, cider, bert-score}. All scores are averages across 4 seeds on the test set. %
For details, see Appendix \ref{sec:app:addres:speak}.

\subsection{Discriminator}
\label{sec:listener}

Our listener is a discriminator model that receives six images in the visual context plus an utterance, and is tasked with identifying the target image that the utterance refers to. To encode the utterance, we use word embeddings trained from scratch to make sure no knowledge leaks from any pretraining. The model combines the visual context and the utterance to produce a multimodal context vector. The listener identifies the target image by comparing this multimodal context vector to the representations of each candidate image via dot-product and selecting the image with the highest score. See Appendix~\ref{sec:app:train_detail:list} for the detailed description of the model architecture.

We train one listener model per domain in Table~\ref{tab:data-stats}.\footnote{We also train a general listener model on all domains which is only used to train the self-aware simulator; see \textsection\ref{sec:sim}.} The models are optimized with Cross Entropy loss using the Adam optimizer. The best models are selected based on resolution accuracy on the validation set. We keep these domain-specific listener models frozen in the rest of the study. See Appendix \ref{sec:app:train_detail:list} for further details. 

\paragraph{Performance} 
We distinguish between \textit{in-domain} (\ind) accuracy---i.e., the resolution accuracy achieved on the test set of the domain on which the listener has been trained---and \textit{out-of-domain} (\ood) accuracy---accuracy on domains the listener has not been exposed to (e.g., the accuracy on images from the \textit{vehicles} domain of a listener exclusively trained on the \textit{food} domain). 
Our listeners are truly domain specific: they are able to identify the target image with an average accuracy of $83.08\%$ in \ind, while their \ood accuracy is  $19.05\%$ on average---barely above a random baseline ($16.67\%$). See Appendix~\ref{sec:app:addres:list} for the full results broken down per domain.

\subsection{Simulator}
\label{sec:sim}

As explained in \textsection\ref{sec:problem}, the speaker is endowed with a simulator module. 
The simulator receives inputs in two parallel streams. 
In one stream, it receives the visual context \viscontext\ coupled with the speaker's planned utterance $\utt_t$, and in the second stream, the visual context along with the language model's initial hidden state \ho. The motivation behind this architectural choice is related to the plug-and-play approach at the core of our proposal. The first stream is inspired by previous work on ToM (e.g., \citealp{rabinowitz2018tom}): its main input is the same as what a listener would receive, an utterance. However, to control generation on the fly, we need to modify the language model's internal representations. Thus, the main reason for the second stream is technical: the gradients from the simulator’s loss cannot flow back to the language model’s hidden states if the input to the simulator is text due to the non-differentiability of the \textit{argmax} operation.\footnote{We observed that using the Gumbel-Softmax trick~\citep{jang2017categorical} led to unstable behaviour.}
The second stream  uses a combination of linear layers and standardization to compute the dot product between \ho\ and \viscontext. The outcomes of the two streams are multiplied to obtain the final representation that is compared to the candidate images. 

We train one audience-aware simulator per domain-specific listener and one self-aware general simulator with Cross Entropy loss using the AdamW optimizer~\cite{loshchilov2017decoupled}.  The training set sizes of both types of simulators are the same, with the target behaviour being different. In the simulation of a general listener, the simulator predicts the behaviour of a listener that was exposed to all domains as the speaker, contrary to one domain in the domain-specific case. We choose the best simulator per listener type based on the simulators' prediction accuracies (more details in Appendix~\ref{sec:app:train_detail:sim}). The simulators are then frozen in the rest of the pipeline.

\paragraph{Performance}
The self-aware simulator achieves an accuracy of $70\%$ when predicting the behaviour of a general listener. The audience-aware simulators predict the behaviour of domain-specific listeners with an average accuracy of $78.20\%$ for \ind samples, and $72.78\%$ for \ood samples.\footnote{Possibly because the general knowledge space is bigger, it could also be more difficult to model a general listener than a domain-specific listener with a limited knowledge space.} The drop in accuracy from \ind samples to \ood samples could be due to difficulties in ascertaining the reactions of a listener on \ood data.
 See details of the results in Appendix \ref{sec:app:addres:sim}.

\section{Audience-Aware Adaptation}
\label{sec:adp_results}

In our framework, adaptation takes place at inference time building on our pretrained, frozen models for the language model, the discriminators and simulators described in \textsection\ref{sec:models}. We first explain our adaptation mechanism (\textsection\ref{sec:adapt_mec}) and then report the results obtained (\textsection\ref{sec:adapt_res}).

\subsection{Adaptation Mechanism}
\label{sec:adapt_mec}

Algorithm \ref{algo:adaptation} describes the adaptation mechanism sketched in \textsection\ref{sec:problem}, which exploits the simulator to iteratively monitor the generation outcomes of the speaker. 
Given the visual context \viscontext, the initial hidden state of the speaker's decoder \ho\ and the currently planned utterance $u_t$, the simulator makes a prediction for the listener's selection.\footnote{To avoid excessive language drift and help regularize utterance generation, at inference time %
we condition \ho\  with the previous gold utterance referring to the target image in the current dialogue (if it exists), as done by \citet{takmaz-etal-2020-refer}.  
This resonates with precautions taken in other plug-and-play approaches against text degeneration~\citep{dathathri2020plug}. \label{ftn:prev_utt}} 
We calculate the Cross Entropy loss between the simulator's prediction and the true target. 
We use the gradients flowing back from this loss to update \ho\ with the Adam optimizer. That is, adaptation is performed by backpropagating the loss to modify only the initial hidden state of the speaker's decoder. Based on the updated \ho, the language model generates a new utterance to be reviewed by the simulator. The mechanism stops when: either (1) the simulator predicts that the listener will choose the gold target image; or (2) when the maximum number of adaptation steps is reached (\siter). At each step, we reset the random seed to ensure that the changes in the sampling of the words are only attributable to the updates to \ho, showing the effects of adaptation directly without being confounded by the stochastic nature of sampling. %

\SetKwInput{KwInput}{Input}                %
\SetKwInput{KwOutput}{Output}              %

\begin{algorithm}[!ht]\small
\DontPrintSemicolon
  
  \KwInput{$\siter:$ maximum number of adaptation steps \\\quad\quad\quad$\lradapt:$ learning rate for adaptation\\\quad\quad\quad$seed:$ random seed}
  \KwData{$\ho:$ speaker's initial hidden state\\ \quad\quad\quad$\viscontext:$ visual context\\
  \quad\quad\quad$\truetarg:$ true target}
  $i \leftarrow 0$\;

   \While{$i \leq \siter$}
   {
        $set\_seed(seed)$ \;
   		$u_t =  Speaker(\viscontext, \ho)$ \;
        $\simout = Simulator(\viscontext, u_t, \ho)$ \;
        $\simpred = \argmax(\simout)$ \;

        \If{$\simpred == \truetarg$}{
        break \;
        }

        $loss = CrossEntropy(\simout, \truetarg)$ \;
        $\ho = backprop(loss, \ho, \lradapt)$ \;
        $i\mathrel{+}=1$ \;
   }
   $t_l = Listener(\viscontext, u_t)$ \;
\caption{Adaptation Mechanism}
\label{algo:adaptation}
\end{algorithm}

\subsection{Results}
\label{sec:adapt_res}

 We evaluate whether our approach leads to increased communicative success, quantified in terms of listener resolution accuracy.
 We report the results for the three settings described in \textsection\ref{sec:problem}. For each of the three modules involved in these settings, we provide an evaluation card~\cite{hupkes2022taxonomy} to clarify the nature of our generalisation tests in Appendix~\ref{sec:appendix-eval-card}.
  
\paragraph{Baseline} 

Table~\ref{tab:list_base_results_5x5} provides a breakdown of resolution accuracies per type of domain-specific listener in the setting without simulation; Table~\ref{tab:adaptive_results_avg} shows the averages. Not surprisingly, the results obtained with generated utterances are lower than those reported in \textsection\ref{sec:listener}. However the patterns are the same: when the speaker agent refers to an image within a domain known to the listener (\ind), the average resolution accuracy is 52.30\%; communication however breaks down in out-of-domain instances, where the average \ood score is 19.06\%, close to random choice.

\begin{table}[ht]\centering
\resizebox{\columnwidth}{!}{
\begin{tabular}{@{}l@{\ \ \ }ccccc}\toprule
\multicolumn{1}{c}{} & \multicolumn{1}{c}{\textbf{app}} & \multicolumn{1}{c}{\textbf{food}} & \multicolumn{1}{c}{\textbf{indoor}} & \multicolumn{1}{c}{\textbf{outdoor}} & \multicolumn{1}{c}{\textbf{vehi}} \\ \midrule
\textbf{appliances}  & \cellcolor{lightgray}$57.61$ & $ 20.10$  & $ 19.92 $ & $ 21.27$ & $ 15.98$\\ 
\textbf{food}  & $ 19.11$ & \cellcolor{lightgray}$54.29$ & $ 18.60$ & $ 18.85$ & $ 18.85$ \\ 
\textbf{indoor}  & $ 22.71$ & $ 19.65$ & \cellcolor{lightgray} $53.62$ & $ 20.82$ & $ 16.77$ \\ 
\textbf{outdoor}  & $ 15.08$ & $ 21.46$ & $ 19.62$ & \cellcolor{lightgray}$52.93$ & $ 17.69$ \\ 
\textbf{vehicles}  & $ 16.36$ & $ 16.17$ & $ 17.41$ & $ 20.13$ & \cellcolor{lightgray}$43.08$ \\ \bottomrule
\end{tabular}
}
\caption{Resolution accuracy in the Baseline setting. Rows indicate the listener domain and columns the evaluation domain. Shaded cells show \ind accuracy. Averages across 5 seeds. Full table with sd's in App.~\ref{sec:app:addres:list}.}
\label{tab:list_base_results_5x5}
\end{table}

\paragraph{Self-aware adaptation}
As shown in Table~\ref{tab:adaptive_results_avg}, with the added capability to simulate and adapt to a generic listener, 
we observe an increase in \ind resolution accuracy (52.30\% vs.\ 65.09\%). Yet, this setting does not help to bridge the knowledge gap between speaker and listener: when the input is \ood for a domain-specific listener, adaptation with a general simulator does not lead to higher communicative success (19.06\% vs.\ 19.11\%).

\begin{table}[ht]%
\resizebox{\columnwidth}{!}{
\begin{tabular}{lccc} \toprule
     & \multicolumn{1}{c}{\bf Baseline} & \multicolumn{1}{c}{\bf Self-aware} & \multicolumn{1}{c}{\bf Audience-aware} \\ \midrule
\ood & $19.06 \pm 0.47$           & $19.11 \pm 1.12$                    & $26.74 \pm 1.48$         \\ 
\ind & $52.30 \pm 1.10$           & $65.09 \pm 1.98$                    & $71.77 \pm 2.16$         \\ \bottomrule
\end{tabular}
}
\caption{Average resolution accuracy for our 3 settings in \ood and \ind. Results on the test set over 5 runs. %
}
\label{tab:adaptive_results_avg}
\end{table}

\paragraph{Audience-aware adaptation}
When the speaker adapts its utterances by predicting the behaviour of a domain-specific listener, we see a significant increase in both \ind and \ood (Table~\ref{tab:adaptive_results_avg}). This indicates that audience-aware adaptation helps in knowledge asymmetric scenarios, including in \ind situations where the agents communicate about a domain known to the listener (65.09\% vs.\ 71.77\%). More importantly, while there is certainly room for improvement, the speaker is able to generate utterances that can more often be resolved in \ood (19\% vs.\ 26.74\%).

\section{Analysis}
\label{sec:analysis}

Our experiments show that simulation-based adaptation leads to more successful communication. In this section, we analyse the speaker model and its generated utterances to understand which neural processing mechanisms and which production strategies are behind our main results.

\subsection{Probing for Domain Information}

We begin with an analysis of the neural representations of the speaker model in the \textit{audience-aware} setting. We focus on $h_0$, the first hidden state of the LSTM decoder. This is the output of the visual encoder on which the simulator module intervenes in order to adapt the speaker's utterance plan. Because $h_0$ is the result of encoding a target image (within a visual context), we expect it to carry information about the semantic domain of the image. If it was not able to differentiate visual domains, it would be very unlikely to successfully adapt to domain-specific listeners. We test this hypothesis using diagnostic probing \cite{adi2017auxiliary,conneau-etal-2018-cram,hupkes2018visualisation}. We train a logistic regression classifier on a 70\% of hidden states $h_0$ collected from the speaker when at test time, and then we assess whether it can predict the image domain corresponding to the remaining 30\% of the hidden states. 
As expected, the probing classifier is able to do so with perfect precision and recall (both equal $1.0$) across the 5 visual domains. Using the same approach, we test whether the domain of the \textit{listener} -- rather than the image domain -- is also encoded in $h_0$.\footnote{
We train a logistic regression classifier on the 70\% split of the $h_0$ but this time using as label the domain of the listener. We then evaluate whether the classifier can predict the listener domain in the 30\% probing test set.}
Our hypothesis is that this should not be the case: before the simulator kicks in, the speaker model has no information on the listener's domain-specific knowledge. Probing accuracy scores vary between $0.13$ and $0.16$ across domains (the random baseline is $0.17$), indicating that indeed the speaker's hidden state does not carry listener information before adaptation.

\begin{figure}
    \centering
    \includegraphics[width=0.8\columnwidth]{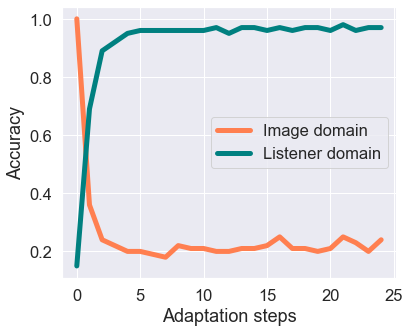}
    \caption{Probing accuracy for image domain and listener domain predictions over adaptation steps. The $0$-th step corresponds to the non-adapted $h_0$.}
    \label{fig:probing_acc}
\end{figure}

As the simulator activates, the original $h_0$ is updated for a maximum of $st_{adp}$ adaptation steps. We now look at the updated hidden states $h_0^1, \ldots, h_0^{st_{adp}}$ and  test whether their encoding of the image and the listener domain changes with adaptation. First, we use the probing classifier previously trained to predict image domains from $h_0$ to test the adapted hidden states. We find that the encoding of the image domain deteriorates with domain-specific adaptation (Figure \ref{fig:probing_acc}). Then, we probe $h_0^1, h_0^2, \ldots$ for listener information and we show that the listener's domain can be predicted almost perfectly from the adapted $h_0$ after only three adaptation steps (Figure \ref{fig:probing_acc}).\footnote{
For this analysis, we train and test one probing classifier for each adaptation step. Using the classifier trained on $h_0$ would not make sense as we showed that it is not possible to extract listener information from non-adapted representations.
}
Taken together, these observations indicate that the neural processing mechanism that leads to more successful interaction is one by which information about the semantic domain of the visual context is replaced by information on the domain of the listener -- and one which only requires a few gradient updates.

\subsection{The Speaker's Adapted Vocabulary}
\label{sec:analysis_vocab}

\begin{figure}
    \centering
    \includegraphics[width=0.8\columnwidth]{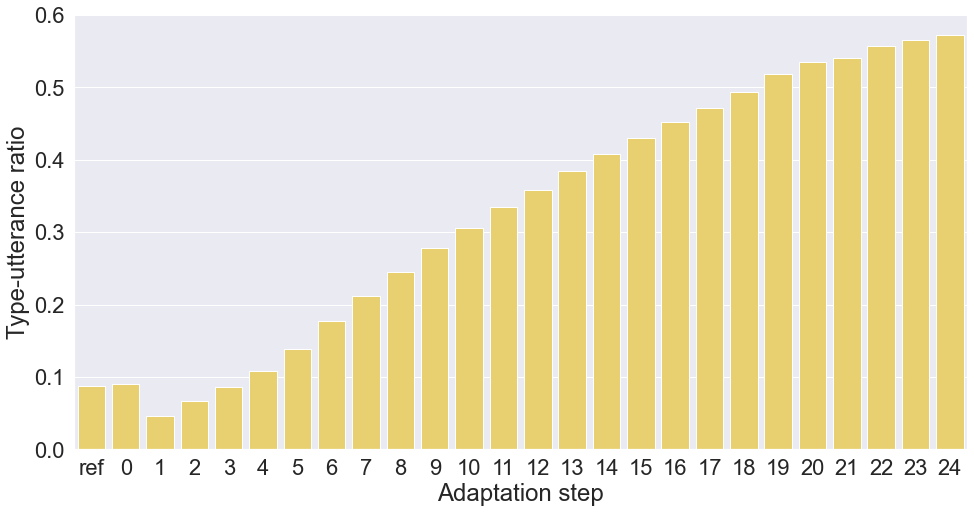}
    \caption{Type-utterance ratio across adaptation steps (type-token ratio in Fig.~\ref{fig:vocab-ttr}, App.~\ref{sec:app:analyses}). Human gold utterances (\textit{ref}) and non-adapted utterances (0) also shown.}
    \label{fig:vocab-utr}
\end{figure}

We analyse macro-level properties of the corpus of adapted utterances as compared to the utterances generated in the simulator-less baseline setting. 
We compute type-utterance ratio and type-token ratio over adaptation steps to monitor the relative size and the variety of the vocabulary as the speaker uses its simulator module. As Figure~\ref{fig:vocab-utr} shows, after an initial drop for the first 1-3 adaptation steps, type-utterance ratio and type-token ratio increase substantially with respect to the non-adapted utterances (and to the gold referring utterances). The speaker vocabulary becomes much more diverse. 
What remains rather stable throughout adaptation, instead, is the unigram part-of-speech distribution (Figure~\ref{fig:pos-ind-ood} in Appendix~\ref{sec:app:analyses}). While, after the first adaptation step, the difference in POS usage is notable (e.g., less punctuation, more nouns), only proper nouns and determiners show substantial changes in relative proportions, with proper nouns increasing and determiners decreasing over time.

\subsection{Adaptation Strategies} 

\begin{figure}
    \centering
    \includegraphics[width=\linewidth]{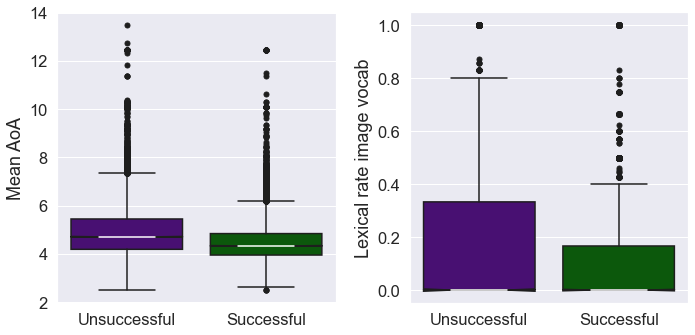}
    \caption{Factors affecting the success of an adapted utterance, age of acquisition (left) and \% of words in an utterance belonging to the target image domain (right).}
    \label{fig:success-plots}
\end{figure}

The trends observed so far characterise the effect of adaptation across steps but they do not differentiate between successful and unsuccessful adaptation. In Figure \ref{fig:success-plots}, we split adapted utterances (the ones actually generated by the speaker when it believed its utterance would be successful) according to whether they lead to a correct listener guess. We observe that more successful utterances contain words with lower age of acquisition\footnote{
Age of Acquisition is a psycholinguistic measure expressing the age at which a word is typically learned. We use the ratings by \citet{kuperman2012age}; they range from 0 to 25.
} (AoA, $t=-28.88$, $p < 0.001$), they show a lower rate of lexical choice from the target image vocabulary ($t=-28.76$, $p < 0.001$), %
and a higher rate of words from the listener vocabulary ($t=5.88$, $p < 0.001$).
The average AoA in an utterance increases with adaptation steps (see Fig.~\ref{fig:aoa-over-steps} in Appendix~\ref{sec:app:analyses}), suggesting that the excessive abstractness of the descriptions may be behind the limited gains we observe with adaptation.

\subsection{Qualitative Inspection}

\begin{figure}
    \centering
    \includegraphics[width=\columnwidth]{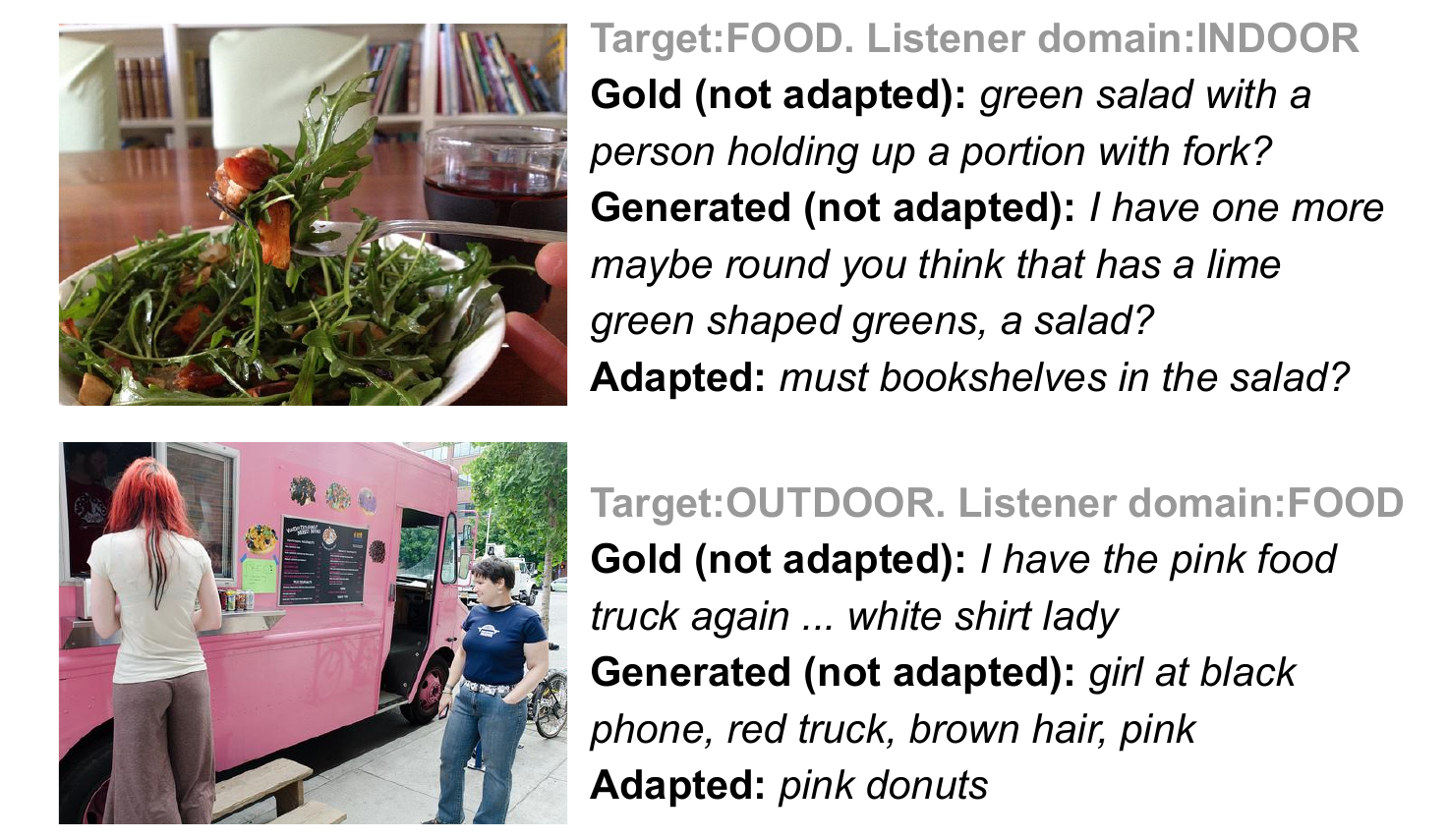}
    \caption{Examples showing how audience-aware adaptation changes the generated utterances. For simplicity, we only show the target images and not the whole visual contexts. We report the final adapted utterances when the adaptation mechanism stops because the simulator predicts that the listener will select the correct image.}
    \label{fig:examples}
\end{figure}

In Figure \ref{fig:examples}, we provide examples of adapted sentences from the test set to demonstrate how the audience-aware adaptation strategies affect the lexical choices made by the language model. In the top example, the image domain is `food'; however, the listener was trained on the `indoor' domain. 
We see that the speaker moves away from generating detailed mentions of food to including a word related to the listener's own domain, \textit{bookshelves}. In the bottom example where the listener has only been exposed to the `food' domain and the image domain is `outdoor', the model avoids mentioning the \textit{truck}. Instead, it produces an utterance containing a prominent color in the image, i.e., pink, and some visible entities that belong to the listener's domain, namely, \textit{donuts}. These observations suggest that the model exploits various adaptation strategies.  

In the whole set of adapted utterances, we observe comprehensible sentences; however, there is also a large number of less fluent, unnatural ones. As we do not use pretrained large language models, sometimes, the speaker's initial utterances themselves are not fluent. The dynamics of adaptation may further exacerbate this situation and lead the language model towards generating unnatural utterances. Such utterances may not be understood by human listeners; yet, they could make sense to artificial listeners. In order to ensure that the adapted utterances are comprehensible to humans, further precautions may be needed, such as incentivizing the generative model to keep the adapted utterances grammatical and fluent, possibly with the aid of human feedback. 

\section{Conclusion}
\label{sec:conclusion}

We focused on a standard reference game---a speaker produces an utterance, and a listener uses it to pick the referent from a visual context. However, our 
setup is \textit{asymmetric}---the speaker has general semantic knowledge, while the listener
has little knowledge of all domains but one (e.g., \emph{food}). Such a setting is a perfect scenario for studying adaptation, i.e., the common process in human communication by which a speaker tunes its language to that of a listener to achieve communicative success. We modeled this mechanism using a \textit{plug-and-play} approach to controllable text generation: the speaker's output is conditioned on the effect of the planned utterance on the listener, as predicted by an internal simulator. Our results show that \emph{speaking the language of a listener} increases communicative success. Through adaptation, the speaker's language becomes less tied to the input domain and more tied to the listener's vocabulary, revealing that audience-aware adaptation can be realized without irreversible changes to generation models.\looseness-1 

Our approach and findings pave the way for pragmatic models that can account for different communicative scenarios. Future work may study adaptation to other dimensions such as age group or sociocultural background. Moreover, adaptation could be explored in multiple `directions'---in our setup, only the speaker adapts. We also simplify the setup by abstracting away the online process that leads to the simulation of the listeners. It would be beneficial to allow the simulators to learn to predict listener behaviour during interaction in an online manner. Finally, our approach could be applied to other and possibly more complex communicative tasks, perhaps in conjunction with a mechanism leveraging human feedback via reinforcement learning.

\section*{Limitations}

Although we use data from dialogues, we do not model collaborative reference, i.e., we do not model continual mutual adaptation. Instead, we focus on the speaker's adaptation to the listener in a single turn, which is certainly a simplified setup. Furthermore, our plug-and-play approach still requires the training of simulators per listener type. However, as we keep the speaker and listener models frozen and use the output obtained from them to train the simulators, this allows us to reduce the required amounts of training. We train the models from scratch using PhotoBook data and do not make use of state-of-the-art large pretrained vision-and-language models that are nowadays commonly based on Transformers, which could be considered a limitation. We opted for this setup as it is more aligned with our research questions, allowing us to control the domain-specificity of the models. We also acknowledge the imbalance in the set sizes of the domains, as well as the possible lexical and visual overlaps in the samples across domains. The overlaps may facilitate the adaptation of certain sentences from one domain to another (asymmetry is not controlled in a fine-grained manner), and this is not uncommon in human communication.

\section*{Ethics Statement}

We are using neither large pretrained language models that have been found to be prone to bias issues nor uncurated data scraped from the internet that would open up myriads of problems. Still, there could be some bias in the PhotoBook data that should be investigated: players might have used offensive or undesirable language in describing images. Therefore, deploying these speakers and listeners directly is not advisable. Our research focusing on the adaptation of a speaker to their audience is done with the aim of improving communicative success within scenarios with knowledge asymmetry following human capabilities of self-monitoring and Theory of Mind. It is possible that adaptation to a specific listener could exacerbate possible biases if the training set of a given listener happens to include more bias. However, the reverse is also the case, where adaptation to underrepresented user groups could be beneficial.

\section*{Acknowledgements}

We would like to thank Arabella Sinclair for her valuable input regarding the conception of this project, Gabriele Gennaro for his contribution to early implementations of the models,  and the members of the Dialogue Modelling Group for their feedback. This research has received funding from the European Research Council (ERC) under the European Union’s Horizon 2020 research and innovation programme (grant agreement No. 819455). Nicolo' Brandizzi carried out the research while visiting the Dialogue Modelling Group in Amsterdam. His contribution was partially supported by the PNRR MUR project PE0000013-FAIR.

\bibliography{adapt}

\begin{thebibliography}{67}
\expandafter\ifx\csname natexlab\endcsname\relax\def\natexlab#1{#1}\fi

\bibitem[{Adi et~al.(2017)Adi, Kermany, Belinkov, Lavi, and
  Goldberg}]{adi2017auxiliary}
Yossi Adi, Einat Kermany, Yonatan Belinkov, Ofer Lavi, and Yoav Goldberg. 2017.
\newblock \href {https://openreview.net/forum?id=BJh6Ztuxl} {Fine-grained
  analysis of sentence embeddings using auxiliary prediction tasks}.
\newblock In \emph{5th International Conference on Learning Representations,
  {ICLR} 2017, Toulon, France, April 24-26, 2017, Conference Track
  Proceedings}. OpenReview.net.

\bibitem[{Alayrac et~al.(2022)Alayrac, Donahue, Luc, Miech, Barr, Hasson, Lenc,
  Mensch, Millican, Reynolds, Ring, Rutherford, Cabi, Han, Gong, Samangooei,
  Monteiro, Menick, Borgeaud, Brock, Nematzadeh, Sharifzadeh, Binkowski,
  Barreira, Vinyals, Zisserman, and Simonyan}]{flamingo}
Jean-Baptiste Alayrac, Jeff Donahue, Pauline Luc, Antoine Miech, Iain Barr,
  Yana Hasson, Karel Lenc, Arthur Mensch, Katie Millican, Malcolm Reynolds,
  Roman Ring, Eliza Rutherford, Serkan Cabi, Tengda Han, Zhitao Gong, Sina
  Samangooei, Marianne Monteiro, Jacob Menick, Sebastian Borgeaud, Andrew
  Brock, Aida Nematzadeh, Sahand Sharifzadeh, Mikolaj Binkowski, Ricardo
  Barreira, Oriol Vinyals, Andrew Zisserman, and Karen Simonyan. 2022.
\newblock \href {https://doi.org/10.48550/ARXIV.2204.14198} {Flamingo: a visual
  language model for few-shot learning}.

\bibitem[{Andreas and Klein(2016)}]{andreas-klein-2016-reasoning}
Jacob Andreas and Dan Klein. 2016.
\newblock \href {https://doi.org/10.18653/v1/D16-1125} {Reasoning about
  pragmatics with neural listeners and speakers}.
\newblock In \emph{Proceedings of the 2016 Conference on Empirical Methods in
  Natural Language Processing}, pages 1173--1182, Austin, Texas. Association
  for Computational Linguistics.

\bibitem[{Bao et~al.(2022)Bao, Ghosh, and Chai}]{bao-etal-2022-learning}
Yuwei Bao, Sayan Ghosh, and Joyce Chai. 2022.
\newblock \href {https://doi.org/10.18653/v1/2022.acl-long.202} {Learning to
  mediate disparities towards pragmatic communication}.
\newblock In \emph{Proceedings of the 60th Annual Meeting of the Association
  for Computational Linguistics (Volume 1: Long Papers)}, pages 2829--2842,
  Dublin, Ireland. Association for Computational Linguistics.

\bibitem[{Ben-David et~al.(2022)Ben-David, Oved, and Reichart}]{pada}
Eyal Ben-David, Nadav Oved, and Roi Reichart. 2022.
\newblock \href {https://doi.org/10.1162/tacl_a_00468} {{PADA}: {E}xample-based
  prompt learning for on-the-fly adaptation to unseen domains}.
\newblock \emph{Transactions of the Association for Computational Linguistics},
  10:414--433.

\bibitem[{Biewald(2020)}]{wandb}
Lukas Biewald. 2020.
\newblock \href {https://www.wandb.com/} {Experiment tracking with weights and
  biases}.
\newblock Software available from wandb.com.

\bibitem[{Brown et~al.(2020)Brown, Mann, Ryder, Subbiah, Kaplan, Dhariwal,
  Neelakantan, Shyam, Sastry, Askell, Agarwal, Herbert-Voss, Krueger, Henighan,
  Child, Ramesh, Ziegler, Wu, Winter, Hesse, Chen, Sigler, Litwin, Gray, Chess,
  Clark, Berner, McCandlish, Radford, Sutskever, and Amodei}]{gpt3}
Tom Brown, Benjamin Mann, Nick Ryder, Melanie Subbiah, Jared~D Kaplan, Prafulla
  Dhariwal, Arvind Neelakantan, Pranav Shyam, Girish Sastry, Amanda Askell,
  Sandhini Agarwal, Ariel Herbert-Voss, Gretchen Krueger, Tom Henighan, Rewon
  Child, Aditya Ramesh, Daniel Ziegler, Jeffrey Wu, Clemens Winter, Chris
  Hesse, Mark Chen, Eric Sigler, Mateusz Litwin, Scott Gray, Benjamin Chess,
  Jack Clark, Christopher Berner, Sam McCandlish, Alec Radford, Ilya Sutskever,
  and Dario Amodei. 2020.
\newblock \href
  {https://proceedings.neurips.cc/paper/2020/file/1457c0d6bfcb4967418bfb8ac142f64a-Paper.pdf}
  {Language models are few-shot learners}.
\newblock In \emph{Advances in Neural Information Processing Systems},
  volume~33, pages 1877--1901. Curran Associates, Inc.

\bibitem[{Clark(1985)}]{clark1985language}
Herbert~H Clark. 1985.
\newblock Language use and language users.
\newblock \emph{Handbook of social psychology (3rd ed.)}, pages 179--231.

\bibitem[{Clark(1996)}]{Clark1996}
Herbert~H. Clark. 1996.
\newblock \emph{Using Language}.
\newblock Cambridge University Press.

\bibitem[{Cohn-Gordon et~al.(2018)Cohn-Gordon, Goodman, and
  Potts}]{cohn-gordon-etal-2018-pragmatically}
Reuben Cohn-Gordon, Noah Goodman, and Christopher Potts. 2018.
\newblock \href {https://doi.org/10.18653/v1/N18-2070} {Pragmatically
  informative image captioning with character-level inference}.
\newblock In \emph{Proceedings of the 2018 Conference of the North {A}merican
  Chapter of the Association for Computational Linguistics: Human Language
  Technologies, Volume 2 (Short Papers)}, pages 439--443, New Orleans,
  Louisiana. Association for Computational Linguistics.

\bibitem[{Conneau et~al.(2018)Conneau, Kruszewski, Lample, Barrault, and
  Baroni}]{conneau-etal-2018-cram}
Alexis Conneau, German Kruszewski, Guillaume Lample, Lo{\"\i}c Barrault, and
  Marco Baroni. 2018.
\newblock \href {https://doi.org/10.18653/v1/P18-1198} {What you can cram into
  a single {\$}{\&}!{\#}* vector: {P}robing sentence embeddings for linguistic
  properties}.
\newblock In \emph{Proceedings of the 56th Annual Meeting of the Association
  for Computational Linguistics (Volume 1: Long Papers)}, pages 2126--2136,
  Melbourne, Australia. Association for Computational Linguistics.

\bibitem[{Corona~Rodriguez et~al.(2019)Corona~Rodriguez, Alaniz, and
  Akata}]{corona2019modeling}
Rodolfo Corona~Rodriguez, Stephan Alaniz, and Zeynep Akata. 2019.
\newblock \href
  {https://proceedings.neurips.cc/paper/2019/file/df308fd90635b28d82558cf580c73ed9-Paper.pdf}
  {{Modeling Conceptual Understanding in Image Reference Games}}.
\newblock In \emph{Advances in Neural Information Processing Systems},
  volume~32. Curran Associates, Inc.

\bibitem[{Dathathri et~al.(2020)Dathathri, Madotto, Lan, Hung, Frank, Molino,
  Yosinski, and Liu}]{dathathri2020plug}
Sumanth Dathathri, Andrea Madotto, Janice Lan, Jane Hung, Eric Frank, Piero
  Molino, Jason Yosinski, and Rosanne Liu. 2020.
\newblock \href {https://openreview.net/forum?id=H1edEyBKDS} {{Plug and Play
  Language Models: A Simple Approach to Controlled Text Generation}}.
\newblock In \emph{International Conference on Learning Representations}.

\bibitem[{Frank and Goodman(2012)}]{frank2012predicting}
Michael~C Frank and Noah~D Goodman. 2012.
\newblock Predicting pragmatic reasoning in language games.
\newblock \emph{Science}, 336(6084):998--998.

\bibitem[{French(1999)}]{FRENCH1999128}
Robert~M. French. 1999.
\newblock \href {https://doi.org/https://doi.org/10.1016/S1364-6613(99)01294-2}
  {Catastrophic forgetting in connectionist networks}.
\newblock \emph{Trends in Cognitive Sciences}, 3(4):128--135.

\bibitem[{Fried et~al.(2021)Fried, Chiu, and Klein}]{fried-etal-2021-reference}
Daniel Fried, Justin Chiu, and Dan Klein. 2021.
\newblock \href {https://doi.org/10.18653/v1/2021.emnlp-main.163}
  {Reference-centric models for grounded collaborative dialogue}.
\newblock In \emph{Proceedings of the 2021 Conference on Empirical Methods in
  Natural Language Processing}, pages 2130--2147, Online and Punta Cana,
  Dominican Republic. Association for Computational Linguistics.

\bibitem[{Goodman and Frank(2016)}]{goodman2016pragmatic}
Noah~D Goodman and Michael~C Frank. 2016.
\newblock Pragmatic language interpretation as probabilistic inference.
\newblock \emph{Trends in cognitive sciences}, 20(11):818--829.

\bibitem[{Goodman and Stuhlm{\"u}ller(2013)}]{goodman2013knowledge}
Noah~D Goodman and Andreas Stuhlm{\"u}ller. 2013.
\newblock Knowledge and implicature: {M}odeling language understanding as
  social cognition.
\newblock \emph{Topics in cognitive science}, 5(1):173--184.

\bibitem[{Greco et~al.(2023)Greco, Bagade, Le, and Bernardi}]{sheadapts}
Claudio Greco, Diksha Bagade, Dieu-Thu Le, and Raffaella Bernardi. 2023.
\newblock \href {https://doi.org/10.3389/frai.2023.1017204} {She adapts to her
  student: An expert pragmatic speaker tailoring her referring expressions to
  the layman listener}.
\newblock \emph{Frontiers in Artificial Intelligence}, 6.

\bibitem[{Haber et~al.(2019)Haber, Baumg{\"a}rtner, Takmaz, Gelderloos, Bruni,
  and Fern{\'a}ndez}]{haber2019photobook}
Janosch Haber, Tim Baumg{\"a}rtner, Ece Takmaz, Lieke Gelderloos, Elia Bruni,
  and Raquel Fern{\'a}ndez. 2019.
\newblock \href {https://www.aclweb.org/anthology/P19-1184.pdf} {The
  {PhotoBook} dataset: Building common ground through visually-grounded
  dialogue}.
\newblock In \emph{Proceedings of the 57th Annual Meeting of the Association
  for Computational Linguistics}, pages 1895--1910.

\bibitem[{Hawkins et~al.(2020)Hawkins, Kwon, Sadigh, and
  Goodman}]{hawkins2020continual}
Robert Hawkins, Minae Kwon, Dorsa Sadigh, and Noah Goodman. 2020.
\newblock \href {https://doi.org/10.18653/v1/2020.conll-1.33} {{{Continual
  Adaptation for Efficient Machine Communication}}}.
\newblock In \emph{Proceedings of the 24th Conference on Computational Natural
  Language Learning}, pages 408--419, Online. Association for Computational
  Linguistics.

\bibitem[{He et~al.(2016)He, Zhang, Ren, and Sun}]{resnet2016}
Kaiming He, Xiangyu Zhang, Shaoqing Ren, and Jian Sun. 2016.
\newblock {Deep Residual Learning for Image Recognition}.
\newblock \emph{2016 IEEE Conference on Computer Vision and Pattern Recognition
  (CVPR)}, pages 770--778.

\bibitem[{Hochreiter and Schmidhuber(1997)}]{lstm}
Sepp Hochreiter and J\"{u}rgen Schmidhuber. 1997.
\newblock \href {https://doi.org/10.1162/neco.1997.9.8.1735} {Long short-term
  memory}.
\newblock \emph{Neural Computation}, 9(8):1735–1780.

\bibitem[{Holtzman et~al.(2020)Holtzman, Buys, Du, Forbes, and
  Choi}]{Holtzman2020The}
Ari Holtzman, Jan Buys, Li~Du, Maxwell Forbes, and Yejin Choi. 2020.
\newblock \href {https://openreview.net/forum?id=rygGQyrFvH} {The curious case
  of neural text degeneration}.
\newblock In \emph{International Conference on Learning Representations}.

\bibitem[{Houlsby et~al.(2019)Houlsby, Giurgiu, Jastrzebski, Morrone,
  De~Laroussilhe, Gesmundo, Attariyan, and Gelly}]{houlsby19a}
Neil Houlsby, Andrei Giurgiu, Stanislaw Jastrzebski, Bruna Morrone, Quentin
  De~Laroussilhe, Andrea Gesmundo, Mona Attariyan, and Sylvain Gelly. 2019.
\newblock \href {https://proceedings.mlr.press/v97/houlsby19a.html}
  {Parameter-efficient transfer learning for {NLP}}.
\newblock In \emph{Proceedings of the 36th International Conference on Machine
  Learning}, volume~97 of \emph{Proceedings of Machine Learning Research},
  pages 2790--2799. PMLR.

\bibitem[{Hupkes et~al.(2022)Hupkes, Giulianelli, Dankers, Artetxe, Elazar,
  Pimentel, Christodoulopoulos, Lasri, Saphra, Sinclair, Ulmer, Schottmann,
  Batsuren, Sun, Sinha, Khalatbari, Ryskina, Frieske, Cotterell, and
  Jin}]{hupkes2022taxonomy}
Dieuwke Hupkes, Mario Giulianelli, Verna Dankers, Mikel Artetxe, Yanai Elazar,
  Tiago Pimentel, Christos Christodoulopoulos, Karim Lasri, Naomi Saphra,
  Arabella Sinclair, Dennis Ulmer, Florian Schottmann, Khuyagbaatar Batsuren,
  Kaiser Sun, Koustuv Sinha, Leila Khalatbari, Maria Ryskina, Rita Frieske,
  Ryan Cotterell, and Zhijing Jin. 2022.
\newblock \href {https://arxiv.org/abs/2210.03050} {State-of-the-art
  generalisation research in {NLP}: {A} taxonomy and review}.
\newblock \emph{CoRR}.

\bibitem[{Hupkes et~al.(2018)Hupkes, Veldhoen, and
  Zuidema}]{hupkes2018visualisation}
Dieuwke Hupkes, Sara Veldhoen, and Willem Zuidema. 2018.
\newblock Visualisation and `diagnostic classifiers' reveal how recurrent and
  recursive neural networks process hierarchical structure.
\newblock \emph{Journal of Artificial Intelligence Research}, 61:907--926.

\bibitem[{Isaacs and Clark(1987)}]{isaacs1987references}
Ellen~A Isaacs and Herbert~H Clark. 1987.
\newblock References in conversation between experts and novices.
\newblock \emph{Journal of experimental psychology: general}, 116(1):26.

\bibitem[{Janarthanam and Lemon(2010)}]{janarthanam2010learning}
Srinivasan Janarthanam and Oliver Lemon. 2010.
\newblock Learning to adapt to unknown users: referring expression generation
  in spoken dialogue systems.
\newblock In \emph{Proceedings of the 48th Annual Meeting of the Association
  for Computational Linguistics}, pages 69--78.

\bibitem[{Jang et~al.(2017)Jang, Gu, and Poole}]{jang2017categorical}
Eric Jang, Shixiang Gu, and Ben Poole. 2017.
\newblock \href {https://openreview.net/forum?id=rkE3y85ee} {Categorical
  reparameterization with gumbel-softmax}.
\newblock In \emph{International Conference on Learning Representations}.

\bibitem[{Keysar(2007)}]{keysar2007communication}
Boaz Keysar. 2007.
\newblock Communication and miscommunication: The role of egocentric processes.

\bibitem[{Kingma and Ba(2015)}]{Adam}
Diederik~P. Kingma and Jimmy Ba. 2015.
\newblock \href {http://arxiv.org/abs/1412.6980} {Adam: {A} method for
  stochastic optimization}.
\newblock In \emph{3rd International Conference on Learning Representations,
  {ICLR} 2015, San Diego, CA, USA, May 7-9, 2015, Conference Track
  Proceedings}.

\bibitem[{Krahmer and van
  Deemter(2012)}]{krahmer-van-deemter-2012-computational}
Emiel Krahmer and Kees van Deemter. 2012.
\newblock \href {https://doi.org/10.1162/COLI_a_00088} {Computational
  generation of referring expressions: A survey}.
\newblock \emph{Computational Linguistics}, 38(1):173--218.

\bibitem[{Kuperman et~al.(2012)Kuperman, Stadthagen-Gonzalez, and
  Brysbaert}]{kuperman2012age}
Victor Kuperman, Hans Stadthagen-Gonzalez, and Marc Brysbaert. 2012.
\newblock Age-of-acquisition ratings for 30,000 {E}nglish words.
\newblock \emph{Behavior Research Methods}, 44(4):978--990.

\bibitem[{Lazaridou et~al.(2020)Lazaridou, Potapenko, and
  Tieleman}]{lazaridou-etal-2020-multi}
Angeliki Lazaridou, Anna Potapenko, and Olivier Tieleman. 2020.
\newblock \href {https://doi.org/10.18653/v1/2020.acl-main.685} {Multi-agent
  communication meets natural language: Synergies between functional and
  structural language learning}.
\newblock In \emph{Proceedings of the 58th Annual Meeting of the Association
  for Computational Linguistics}, pages 7663--7674, Online. Association for
  Computational Linguistics.

\bibitem[{Levelt(1993)}]{levelt1993speaking}
Willem~J.M. Levelt. 1993.
\newblock \emph{Speaking: From intention to articulation}.
\newblock MIT press.

\bibitem[{Li and Liang(2021)}]{prefix}
Xiang~Lisa Li and Percy Liang. 2021.
\newblock \href {https://doi.org/10.18653/v1/2021.acl-long.353} {Prefix-tuning:
  Optimizing continuous prompts for generation}.
\newblock In \emph{Proceedings of the 59th Annual Meeting of the Association
  for Computational Linguistics and the 11th International Joint Conference on
  Natural Language Processing (Volume 1: Long Papers)}, pages 4582--4597,
  Online. Association for Computational Linguistics.

\bibitem[{Lin(2004)}]{Lin2004}
Chin-Yew Lin. 2004.
\newblock \href {https://www.aclweb.org/anthology/W04-1013} {{ROUGE}: A package
  for automatic evaluation of summaries}.
\newblock In \emph{Text Summarization Branches Out}, pages 74--81, Barcelona,
  Spain. Association for Computational Linguistics.

\bibitem[{Lin et~al.(2014)Lin, Maire, Belongie, Hays, Perona, Ramanan,
  Doll{\'a}r, and Zitnick}]{lin2014coco}
Tsung-Yi Lin, Michael Maire, Serge Belongie, James Hays, Pietro Perona, Deva
  Ramanan, Piotr Doll{\'a}r, and C~Lawrence Zitnick. 2014.
\newblock {Microsoft {COCO}: {C}ommon {O}bjects in {C}ontext}.
\newblock In \emph{European conference on computer vision}, pages 740--755.
  Springer.

\bibitem[{Liu et~al.(2016)Liu, Lowe, Serban, Noseworthy, Charlin, and
  Pineau}]{liu-etal-2016-evaluate}
Chia-Wei Liu, Ryan Lowe, Iulian Serban, Mike Noseworthy, Laurent Charlin, and
  Joelle Pineau. 2016.
\newblock \href {https://doi.org/10.18653/v1/D16-1230} {How {NOT} to evaluate
  your dialogue system: An empirical study of unsupervised evaluation metrics
  for dialogue response generation}.
\newblock In \emph{Proceedings of the 2016 Conference on Empirical Methods in
  Natural Language Processing}, pages 2122--2132, Austin, Texas. Association
  for Computational Linguistics.

\bibitem[{Loshchilov and Hutter(2017)}]{loshchilov2017decoupled}
Ilya Loshchilov and Frank Hutter. 2017.
\newblock Decoupled weight decay regularization.
\newblock \emph{arXiv preprint arXiv:1711.05101}.

\bibitem[{Mao et~al.(2016)Mao, Huang, Toshev, Camburu, Yuille, and
  Murphy}]{mao2016generation}
Junhua Mao, Jonathan Huang, Alexander Toshev, Oana Camburu, Alan~L Yuille, and
  Kevin Murphy. 2016.
\newblock Generation and comprehension of unambiguous object descriptions.
\newblock In \emph{Proceedings of the IEEE conference on computer vision and
  pattern recognition}, pages 11--20.

\bibitem[{McCloskey and Cohen(1989)}]{MCCLOSKEY1989109}
Michael McCloskey and Neal~J. Cohen. 1989.
\newblock \href {https://doi.org/https://doi.org/10.1016/S0079-7421(08)60536-8}
  {Catastrophic interference in connectionist networks: The sequential learning
  problem}.
\newblock volume~24 of \emph{Psychology of Learning and Motivation}, pages
  109--165. Academic Press.

\bibitem[{Monroe et~al.(2017)Monroe, Hawkins, Goodman, and
  Potts}]{monroe-etal-2017-colors}
Will Monroe, Robert~X.D. Hawkins, Noah~D. Goodman, and Christopher Potts. 2017.
\newblock \href {https://doi.org/10.1162/tacl_a_00064} {Colors in context: A
  pragmatic neural model for grounded language understanding}.
\newblock \emph{Transactions of the Association for Computational Linguistics},
  5:325--338.

\bibitem[{Ohashi and Higashinaka(2022)}]{ohashi-higashinaka-2022-adaptive}
Atsumoto Ohashi and Ryuichiro Higashinaka. 2022.
\newblock \href {https://aclanthology.org/2022.coling-1.19} {Adaptive natural
  language generation for task-oriented dialogue via reinforcement learning}.
\newblock In \emph{Proceedings of the 29th International Conference on
  Computational Linguistics}, pages 242--252, Gyeongju, Republic of Korea.
  International Committee on Computational Linguistics.

\bibitem[{Papineni et~al.(2002)Papineni, Roukos, Ward, and Zhu}]{Papineni:2002}
Kishore Papineni, Salim Roukos, Todd Ward, and Wei-Jing Zhu. 2002.
\newblock \href {https://doi.org/10.3115/1073083.1073135} {{BLEU}: A method for
  automatic evaluation of machine translation}.
\newblock In \emph{Proceedings of the 40th Annual Meeting on Association for
  Computational Linguistics}, pages 311--318. Association for Computational
  Linguistics.

\bibitem[{Pascual et~al.(2021)Pascual, Egressy, Meister, Cotterell, and
  Wattenhofer}]{pascual-etal-2021-plug-play}
Damian Pascual, Beni Egressy, Clara Meister, Ryan Cotterell, and Roger
  Wattenhofer. 2021.
\newblock \href {https://doi.org/10.18653/v1/2021.findings-emnlp.334} {A
  plug-and-play method for controlled text generation}.
\newblock In \emph{Findings of the Association for Computational Linguistics:
  EMNLP 2021}, pages 3973--3997, Punta Cana, Dominican Republic. Association
  for Computational Linguistics.

\bibitem[{Pfeiffer et~al.(2020{\natexlab{a}})Pfeiffer, R{\"u}ckl{\'e}, Poth,
  Kamath, Vuli{\'c}, Ruder, Cho, and Gurevych}]{adapterhub}
Jonas Pfeiffer, Andreas R{\"u}ckl{\'e}, Clifton Poth, Aishwarya Kamath, Ivan
  Vuli{\'c}, Sebastian Ruder, Kyunghyun Cho, and Iryna Gurevych.
  2020{\natexlab{a}}.
\newblock \href {https://doi.org/10.18653/v1/2020.emnlp-demos.7}
  {{A}dapter{H}ub: A framework for adapting transformers}.
\newblock In \emph{Proceedings of the 2020 Conference on Empirical Methods in
  Natural Language Processing: System Demonstrations}, pages 46--54, Online.
  Association for Computational Linguistics.

\bibitem[{Pfeiffer et~al.(2020{\natexlab{b}})Pfeiffer, Vuli{\'c}, Gurevych, and
  Ruder}]{madx}
Jonas Pfeiffer, Ivan Vuli{\'c}, Iryna Gurevych, and Sebastian Ruder.
  2020{\natexlab{b}}.
\newblock \href {https://doi.org/10.18653/v1/2020.emnlp-main.617} {{MAD-X}:
  {A}n {A}dapter-{B}ased {F}ramework for {M}ulti-{T}ask {C}ross-{L}ingual
  {T}ransfer}.
\newblock In \emph{Proceedings of the 2020 Conference on Empirical Methods in
  Natural Language Processing (EMNLP)}, pages 7654--7673, Online. Association
  for Computational Linguistics.

\bibitem[{Pickering and Garrod(2004)}]{Pickering2004-PICTAM}
Martin~J. Pickering and Simon Garrod. 2004.
\newblock Toward a mechanistic psychology of dialogue.
\newblock \emph{Behavioral and Brain Sciences}, 27(2):169--190.

\bibitem[{Pickering and Garrod(2013)}]{pickering2013integrated}
Martin~J Pickering and Simon Garrod. 2013.
\newblock An integrated theory of language production and comprehension.
\newblock \emph{Behavioral and brain sciences}, 36(4):329--347.

\bibitem[{Premack and Woodruff(1978)}]{premack1978tom}
David Premack and Guy Woodruff. 1978.
\newblock Does the chimpanzee have a theory of mind?
\newblock \emph{Behavioral and Brain Sciences}, 1(4):515--526.

\bibitem[{Qin et~al.(2022)Qin, Welleck, Khashabi, and Choi}]{qin2022cold}
Lianhui Qin, Sean Welleck, Daniel Khashabi, and Yejin Choi. 2022.
\newblock Cold decoding: Energy-based constrained text generation with langevin
  dynamics.
\newblock In \emph{Proceedings of the 36th Conference on Neural Information
  Processing Systems (NeurIPS 2022)}.

\bibitem[{Rabinowitz et~al.(2018{\natexlab{a}})Rabinowitz, Perbet, Song, Zhang,
  Eslami, and Botvinick}]{rabinowitz2018tom}
Neil Rabinowitz, Frank Perbet, Francis Song, Chiyuan Zhang, S.~M.~Ali Eslami,
  and Matthew Botvinick. 2018{\natexlab{a}}.
\newblock \href {http://proceedings.mlr.press/v80/rabinowitz18a.html} {{Machine
  Theory of Mind}}.
\newblock In \emph{Proceedings of the 35th International Conference on Machine
  Learning}, volume~80 of \emph{Proceedings of Machine Learning Research},
  pages 4218--4227. PMLR.

\bibitem[{Rabinowitz et~al.(2018{\natexlab{b}})Rabinowitz, Perbet, Song, Zhang,
  Eslami, and Botvinick}]{abs-1802-07740}
Neil~C. Rabinowitz, Frank Perbet, H.~Francis Song, Chiyuan Zhang, S.~M.~Ali
  Eslami, and Matthew~M. Botvinick. 2018{\natexlab{b}}.
\newblock \href {http://arxiv.org/abs/1802.07740} {Machine theory of mind}.
\newblock \emph{CoRR}, abs/1802.07740.

\bibitem[{Reiter and Dale(1997)}]{reiter1997building}
Ehud Reiter and Robert Dale. 1997.
\newblock Building applied natural language generation systems.
\newblock \emph{Natural Language Engineering}, 3(1):57--87.

\bibitem[{Roelofs(2020)}]{roelofs2020self}
Ardi Roelofs. 2020.
\newblock Self-monitoring in speaking: In defense of a comprehension-based
  account.
\newblock \emph{Journal of Cognition}, 3(1).

\bibitem[{Saxton(2009)}]{saxton2009inevitability}
Matthew Saxton. 2009.
\newblock The inevitability of child directed speech.
\newblock In \emph{Language acquisition}, pages 62--86. Springer.

\bibitem[{Singh et~al.(2023)Singh, Ding, Saxe, Hill, and
  Lampinen}]{singh-etal-2023-know}
Aaditya~K Singh, David Ding, Andrew Saxe, Felix Hill, and Andrew Lampinen.
  2023.
\newblock \href {https://aclanthology.org/2023.eacl-main.279} {Know your
  audience: specializing grounded language models with listener subtraction}.
\newblock In \emph{Proceedings of the 17th Conference of the European Chapter
  of the Association for Computational Linguistics}, pages 3884--3911,
  Dubrovnik, Croatia. Association for Computational Linguistics.

\bibitem[{Takmaz et~al.(2020)Takmaz, Giulianelli, Pezzelle, Sinclair, and
  Fern{\'a}ndez}]{takmaz-etal-2020-refer}
Ece Takmaz, Mario Giulianelli, Sandro Pezzelle, Arabella Sinclair, and Raquel
  Fern{\'a}ndez. 2020.
\newblock \href {https://doi.org/10.18653/v1/2020.emnlp-main.353} {{R}efer,
  {R}euse, {R}educe: {G}enerating {S}ubsequent {R}eferences in {V}isual and
  {C}onversational {C}ontexts}.
\newblock In \emph{Proceedings of the 2020 Conference on Empirical Methods in
  Natural Language Processing (EMNLP)}, pages 4350--4368, Online. Association
  for Computational Linguistics.

\bibitem[{Tomasello(2005)}]{tomasello2005constructing}
Michael Tomasello. 2005.
\newblock \emph{Constructing a language: {A} usage-based theory of language
  acquisition}.
\newblock Harvard university press.

\bibitem[{Vedantam et~al.(2017)Vedantam, Bengio, Murphy, Parikh, and
  Chechik}]{vedantam2017context}
Ramakrishna Vedantam, Samy Bengio, Kevin Murphy, Devi Parikh, and Gal Chechik.
  2017.
\newblock Context-aware captions from context-agnostic supervision.
\newblock In \emph{Proceedings of the IEEE Conference on Computer Vision and
  Pattern Recognition}, pages 251--260.

\bibitem[{Vedantam et~al.(2015)Vedantam, Lawrence~Zitnick, and Parikh}]{cider}
Ramakrishna Vedantam, C~Lawrence~Zitnick, and Devi Parikh. 2015.
\newblock \href
  {https://www.cv-foundation.org/openaccess/content_cvpr_2015/papers/Vedantam_CIDEr_Consensus-Based_Image_2015_CVPR_paper.pdf}
  {Cider: Consensus-based image description evaluation}.
\newblock In \emph{Proceedings of the IEEE conference on Computer Vision and
  Pattern Recognition (CVPR)}, pages 4566--4575.

\bibitem[{Yu et~al.(2017)Yu, Tan, Bansal, and Berg}]{Yu2017AJS}
Licheng Yu, Hao Tan, Mohit Bansal, and Tamara~L. Berg. 2017.
\newblock A joint speaker-listener-reinforcer model for referring expressions.
\newblock \emph{2017 IEEE Conference on Computer Vision and Pattern Recognition
  (CVPR)}, pages 3521--3529.

\bibitem[{Zarrie{\ss} and Schlangen(2019)}]{zarriess-schlangen-2019-know}
Sina Zarrie{\ss} and David Schlangen. 2019.
\newblock \href {https://doi.org/10.18653/v1/P19-1063} {Know what you don{'}t
  know: Modeling a pragmatic speaker that refers to objects of unknown
  categories}.
\newblock In \emph{Proceedings of the 57th Annual Meeting of the Association
  for Computational Linguistics}, pages 654--659, Florence, Italy. Association
  for Computational Linguistics.

\bibitem[{Zhang et~al.(2020)Zhang, Kishore, Wu, Weinberger, and
  Artzi}]{bert-score}
Tianyi Zhang, Varsha Kishore, Felix Wu, Kilian~Q. Weinberger, and Yoav Artzi.
  2020.
\newblock \href {https://openreview.net/forum?id=SkeHuCVFDr} {{BERTScore:
  Evaluating Text Generation with BERT}}.
\newblock In \emph{International Conference on Learning Representations}.

\bibitem[{Zhu et~al.(2021)Zhu, Neubig, and Bisk}]{zhu2021few}
Hao Zhu, Graham Neubig, and Yonatan Bisk. 2021.
\newblock Few-shot language coordination by modeling theory of mind.
\newblock In \emph{International Conference on Machine Learning}, pages
  12901--12911. PMLR.

\end{thebibliography}
\bibliographystyle{acl_natbib}

\clearpage

\appendix

\section*{Appendix}

\section{Training Details}
\label{sec:app:train_detail}

We provide the details of the setups of the generative language model in \textsection\ref{sec:app:train_detail:speak}, the discriminators in \textsection\ref{sec:app:train_detail:list}, the simulators in \textsection\ref{sec:app:train_detail:sim} and the adaptation mechanism in \textsection\ref{sec:app:train_detail:adapt_pipeline}. 
We use Python version 3.9.0 and PyTorch version 1.11.0 in the development and testing of all our models. In Table \ref{tab:hyper-param}, we report the hyperparameters used in the training of our final models. %

\begin{table}[h]
\centering
\begin{tabular}{|l|ccc|}
\hline
              & LM & Disc & Simulator \\ \hline
Learning rate & 0.0001  & 0.0001   & 0.0004    \\ \hline
Batch size    & 3       & 64       & 32        \\ \hline
Dropout       & 0.3     & 0.2      & 0         \\ \hline
Attention dim & 512     & 512      & 1024      \\ \hline
Embed dim     & 1024    & 768      & 1024      \\ \hline
Hidden dim    & 512     & 512      & 1024      \\ \hline
Patience      & 30      & 30       & 5         \\ \hline
\end{tabular}
\caption{Hyperparameters used for training the generative language model, discriminator and simulator models.}
\label{tab:hyper-param}
\end{table}

\subsection{Generative Language Model} 
\label{sec:app:train_detail:speak}

In addition to the main hyperparameters listed in Table \ref{tab:hyper-param}, the language model requires several additional parameters. In nucleus sampling, we set the \textit{p} value for \textit{top-p} to 0.9 and sample from a vocabulary that consists of the words in the training splits of all 5 domains. The maximum length of the generated utterances is set to 30. The model is initialized and trained with 4 different seeds, which yield similar performances. We use an early stopping patience of 30 epochs based on the validation set scores.\footnote{We use the `nlg-eval' library at \url{https://github.com/Maluuba/nlg-eval} to obtain scores for the common NLG metrics and also use BERTScore version 0.3.11 provided at \url{https://github.com/Tiiiger/bert_score}.}

Regarding the architectural details of the visually-conditioned language model, in the visual encoder we feed both the standardized %
target image vector and the concatenation of the six images in the full visual context %
into a linear layer followed by the ReLU non-linearity. We then concatenate the ensuing representations of the target image with the visual context and once more apply a linear layer followed by a ReLU non-linearity to obtain the final visual context, \viscontext. %
This visual context is used to initialize a bidirectional LSTM encoder that takes as input the previous utterance referring to the target image in the current dialogue, if exists (see  footnote~\ref{ftn:prev_utt}), otherwise a special token indicating the absence of such an utterance. The final forward and backward hidden states of this encoder are concatenated, go through a linear layer and $tanh$ non-linearity. The output is then set as the initial hidden state \ho\ of the LSTM decoder~\citep{lstm}.

\subsection{Discriminators}
\label{sec:app:train_detail:list}
In these models instantiating the listeners, the word embeddings go through a dropout layer and a linear layer followed by the Leaky-ReLU non-linearity, after which standardization is applied. The visual context is processed in the same way as in the generative language model. Each word representation is concatenated with the representation of the visual context. The resulting vectors go through a linear layer and ReLU. Finally, we apply attention over these vectors to obtain the attention-weighted multimodal context vector. It is this context vector that is compared to the representations of candidate images via dot product.

We use the same set of hyperparameters for each domain as shown in Table \ref{tab:hyper-param}. The domain-specific listener models were selected based on their accuracy on the in-domain validation set. We report accuracy and MRR on the in-domain and out-of-domain test sets in Table \ref{tab:list_train_domain_golden}.

\paragraph{\ood word masking} 
Our listeners are initialized with the same vocabulary comprising all the words in the training data. However, the domain-specific listeners only learn the words that exist in their own training sets. Therefore, if the speaker generates an \ood word for a domain-specific listener, in order not to further confound the effects of adaptation on the listeners, we mask the word with the $<$unk$>$ vector. This vector is the same across all domains.

\subsection{Simulator}
\label{sec:app:train_detail:sim}
We select the simulator models based on their accuracy in predicting the behaviour of the listener models on the validation set. %
The simulator models are trained using the AdamW optimizer \cite{loshchilov2017decoupled} with a weight decay of 0.0001, and a plateau learning scheduler with a patience of 2, a factor of 0.5, a threshold of 0.5. %

\subsection{Adaptation Mechanism}
\label{sec:app:train_detail:adapt_pipeline}
We optimize the values of the number of adaptation steps and the learning rate for the adaptation mechanism. We perform 2 hyperparameter sweeps using the Weight \& Biases (WandB) platform~\cite{wandb}, evaluating a range of values. We find a positive correlation between both hyperparameters and adaptation accuracy, with Pearson's correlation coefficients of $0.71$ for the learning rate, and $0.66$ for the number of steps.

\section{Additional Results}
\label{sec:app:addres}

Here, we provide additional results yielded by our models for the speaker in \textsection\ref{sec:app:addres:speak}, the listener in \textsection\ref{sec:app:addres:list}, the simulator in \textsection\ref{sec:app:addres:sim} and for the adaptation mechanism in \textsection\ref{sec:app:addres:adapt}. 

\subsection{Speaker Results}
\label{sec:app:addres:speak}

We provide the detailed results of the speaker model on the test set in Table~\ref{tab:speak_results} with the averages and standard deviations over 4 runs.

\begin{table*}[ht!]
\centering
\begin{tabular}{|ccccc|}
\hline
\textbf{BLEU-1} & \textbf{BLEU-2} & \textbf{BLEU-3} & \textbf{BLEU-4} & \textbf{Rouge} \\ \hline
$40.06 \pm 1.60$ & $23.81 \pm 1.51$ & $14.09 \pm 1.20$ & $8.46 \pm 0.89$ & $32.92 \pm 0.93$ \\ \hline
\textbf{CIDEr} & \textbf{BertScore - Recall} & \textbf{BertScore-Recall - F1} & \textbf{BertScore - Precision} &  \\ \hline
$44.07 \pm 1.68$ & $58.91 \pm 0.19$ & $57.7 \pm 0.12$ & $57.9 \pm 0.16$ &  \\ \hline
\end{tabular}
\caption{Speaker results on the test set as measured by common natural language generation evaluation metrics.}
 \label{tab:speak_results}
\end{table*}

\subsection{Listener Results}
\label{sec:app:addres:list}
Table \ref{tab:list_train_domain_golden} reports the domain-specific listener performances on \ind and \ood gold data. We observe that the domain-specific listeners perform well in in-domain settings and perform close to the random baseline in \ood settings.

Table \ref{tab:list_baseline_detailed} presents the domain-specific listener accuracies on speaker-generated input. Especially in \ind settings, we see lower scores as compared to use of the gold data, presumably because the listener models were trained on gold data.

\begin{table*}[ht!]
\centering
\begin{tabular}{|l|c|cc|cc|}
\hline
\multirow{2}{*}{\textbf{Domain}} & \multirow{2}{*}{\textbf{Epoch}} & \multicolumn{2}{c}{\ind} & \multicolumn{2}{c|}{\ood} \\ \cline{3-6} 
 &  & \multicolumn{1}{c}{\textbf{Accuracy}} & \multicolumn{1}{c}{\textbf{MRR}} & \multicolumn{1}{c}{\textbf{Accuracy}} & \multicolumn{1}{c|}{\textbf{MRR}} \\ \hline
\textbf{Appliances} & 23 & $ 84.12 \pm 0.33$ & $90.27 \pm 0.10$ & $20.28 \pm 0.23$ & $44.07 \pm 0.11$ \\ \hline
\textbf{Food} & 21 & $85.40 \pm 0.28$ & $91.20 \pm 0.20$ & $17.72 \pm 0.18$ & $42.42 \pm 0.06$ \\ \hline
\textbf{Indoor} & 14 & $82.94 \pm 0.13$ & $89.32 \pm 0.09$ & $19.14 \pm 0.09$ & $43.46 \pm 0.06$ \\ \hline
\textbf{Outdoor} & 19 & $83.96 \pm 0.23$ & $90.01 \pm 0.14$ & $19.64 \pm 0.07$ & $43.52 \pm 0.06$ \\ \hline
\textbf{Vehicles} & 17 & $78.99 \pm 0.35$ & $86.81 \pm 0.14$ & $18.46 \pm 0.28$ & $42.36 \pm 0.20$ \\ \hline
\end{tabular}
\caption{Listener performance on gold utterances. Accuracy and MRR for the in-domain (\ind) and out-of-domain (\ood) samples given to listeners trained on specific domains (indicated under the `Domain' column).}
\label{tab:list_train_domain_golden}
\end{table*}

\begin{table*}[ht!]\centering
\begin{tabular}{|l|ccccc|}
\hline
\multicolumn{1}{|c|}{\multirow{2}{*}{\textbf{\begin{tabular}[c]{@{}c@{}}Listener\\ domain\end{tabular}}}} & \multicolumn{5}{c|}{\textbf{Data domain}} \\ %
\multicolumn{1}{|c|}{} & \multicolumn{1}{c}{\textbf{appliances}} & \multicolumn{1}{c}{\textbf{food}} & \multicolumn{1}{c}{\textbf{indoor}} & \multicolumn{1}{c}{\textbf{outdoor}} & \multicolumn{1}{c|}{\textbf{vehicles}} \\ \hline
\textbf{appliances}  & $ 57.61 \pm 1.38 $ & $ 20.10 \pm 0.63 $ & $ 19.92 \pm 0.47 $ & $ 21.27 \pm 0.83 $ & $ 15.98 \pm 0.82 $\\ \hline
\textbf{food}  & $ 19.11 \pm 1.70 $ & $ 54.29 \pm 1.06 $ & $ 18.60 \pm 0.84 $ & $ 18.85 \pm 0.49 $ & $ 18.85 \pm 0.49 $ \\ \hline
\textbf{indoor}  & $ 22.71 \pm 1.30 $ & $ 19.65 \pm 1.77 $ & $ 53.62 \pm 0.79 $ & $ 20.82 \pm 1.05 $ & $ 16.77 \pm 0.79 $ \\ \hline
\textbf{outdoor}  & $ 15.08 \pm 1.04 $ & $ 21.46 \pm 0.70 $ & $ 19.62 \pm 0.69 $ & $ 52.93 \pm 1.11 $ & $ 17.69 \pm 0.97 $ \\ \hline
\textbf{vehicles}  & $ 16.36 \pm 1.55 $ & $ 16.17 \pm  0.81 $ & $ 17.41 \pm 0.64 $ & $ 20.13 \pm 0.59 $ & $ 43.08 \pm 1.16 $ \\ \hline
\end{tabular}
\caption{Listener accuracies on speaker-generated data. Each row indicates the domain a listener was trained on and the columns indicate the domain of the input samples. Results over 5 seeds.}
\label{tab:list_baseline_detailed}
\end{table*}

\subsection{Simulator Results}
\label{sec:app:addres:sim}
The detailed outcomes of the simulator models are reported in Table \ref{tab:sim_train_avg}. Here, we also report the results for the subset where the \emph{listener} made a correct prediction (Pos) vs. it made an incorrect prediction (Neg). The simulators are better able to capture the correct listener behaviour, possibly because during the training of simulators, in-domain data provides a clear picture of listener's correct behaviour.

\begin{table*}[ht]
\centering
\begin{tabular}{|c|c|ccc|}
\hline
\textbf{Simulator} & \textbf{Setting} & \textbf{Avg} & \textbf{Pos} & \textbf{Neg} \\ \hline
all domains & \multicolumn{1}{c|}{--} & \multicolumn{1}{c}{$69.97 \pm 0.79$} & \multicolumn{1}{c}{$85.15 \pm 1.39$} & \multicolumn{1}{c|}{$54.73 \pm 0.76$}  \\ \hline
\multirow{2}{*}{domain specific} & \ind & $78.20 \pm 1.26$ & $88.09 \pm 1.98$ & $67.36 \pm 2.96$  \\ %
 & \ood & $72.78 \pm 0.56$ & $73.67 \pm 1.69$ & $72.58 \pm 0.71$ \\ \hline
\end{tabular}
\caption{Simulator's accuracy in predicting the behaviour of a listener knowledgeable about \textit{all domains} (as the speaker) and a listener with \textit{domain-specific} knowledge for \ind and \ood samples. `Avg' is the overall accuracy, `Pos' and `Neg' are the percentages of correct predictions for the samples where the listener picked the correct (Pos) and the incorrect image (Neg).} %
\label{tab:sim_train_avg}
\end{table*}

\subsection{Adaptation Results}
\label{sec:app:addres:adapt}

In Table \ref{tab:adaptive_results_domain_GOA}, we provide the test set results of the adaptation pipeline, broken down into domains and for \ind and \ood inputs separately. The outcomes show that adaptation has effects in both \ind and \ood settings, increasing resolution accuracies over speaker-generated utterances.

\begin{table*}[]
\centering
\begin{tabular}{|l|ccc|ccc|}
\hline
                    & \multicolumn{3}{c|}{\textbf{OOD}}                      & \multicolumn{3}{c|}{\textbf{IND}}                      \\ \hline
                    & \textbf{Golden} & \textbf{Speaker} & \textbf{Adapted} & \textbf{Golden} & \textbf{Speaker} & \textbf{Adapted} \\ \hline
\textbf{appliances} & 20.21         & 19.30           & 27.74          & 84.21         & 57.21           & 74.28          \\ \hline
\textbf{indoor}     & 18.50         & 19.53           & 28.34          & 83.22         & 52.94           & 69.62          \\ \hline
\textbf{food}       & 17.06         & 18.31           & 26.26          & 85.61         & 55.54           & 78.15          \\ \hline
\textbf{outdoor}    & 18.89         & 18.54           & 26.21          & 84.38         & 52.83           & 73.04          \\ \hline
\textbf{vehicles}   & 18.25         & 17.35           & 25.16          & 78.67         & 42.09           & 63.75          \\ \hline
\end{tabular}
\caption{Test results for the audience-aware adaptation pipeline, 5 seeds for each domain.}
\label{tab:adaptive_results_domain_GOA}
\end{table*}

\section{Evaluation Cards}
\label{sec:appendix-eval-card}
For each of the three main modules in our experiments, we provide an evaluation card
to clarify the nature of our generalisation tests.\footnote{
    \url{https://genbench.org/eval_cards}
} See Table~\ref{tab:eval-card-generator} for the generator, Table~\ref{tab:eval-card-simulator} for the simulator, and Table~\ref{tab:eval-card-listener} for the listener.
We also register our work in the GenBench evolving survey of generalisation in NLP~\cite{hupkes2022taxonomy}.\footnote{
    \url{https://genbench.org/references}
}\looseness-1

\newcommand{\tabularwidth}{\columnwidth}

\newcommand{\expone}{$\square$}
\newcommand{\exptwo}{$\bigtriangleup$}
\newcommand{\expthree}{$\bigcirc$}
        
\renewcommand{\arraystretch}{1.1}         
\setlength{\tabcolsep}{0mm}    

\begin{table}[]
\centering
\begin{tabular}{|p{\tabularwidth}<{\centering}|}         
\hline
               
\rowcolor{gray!60}               
\textbf{Motivation} \\               
\footnotesize
\begin{tabular}{p{0.25\tabularwidth}<{\centering} p{0.25\tabularwidth}<{\centering} p{0.25\tabularwidth}<{\centering} p{0.25\tabularwidth}<{\centering}}                        
\textit{Practical} & \textit{Cognitive} & \textit{Intrinsic} & \textit{Fairness}\\
\expone\hspace{0.8mm}\exptwo\hspace{0.8mm}\expthree\hspace{0.8mm}		%
& \expone\hspace{0.8mm}\exptwo\hspace{0.8mm}\expthree\hspace{0.8mm}		%
& 		%
& 		%

\vspace{2mm} \\
\end{tabular}\\
               
\rowcolor{gray!60}               
\textbf{Generalisation type} \\               
\footnotesize
\begin{tabular}{m{0.17\tabularwidth}<{\centering} m{0.20\tabularwidth}<{\centering} m{0.14\tabularwidth}<{\centering} m{0.17\tabularwidth}<{\centering} m{0.18\tabularwidth}<{\centering} m{0.14\tabularwidth}<{\centering}}                   
\textit{Compo- sitional} & \textit{Structural} & \textit{Cross Task} & \textit{Cross Language} & \textit{Cross Domain} & \textit{Robust- ness}\\
& 		%
& 		%
& 		%
& \expone\hspace{0.8mm}\exptwo\hspace{0.8mm}\expthree\hspace{0.8mm}		%
& 		%

\vspace{2mm} \\
\end{tabular}\\
             
\rowcolor{gray!60}             
\textbf{Shift type} \\             
\footnotesize
\begin{tabular}{p{0.25\tabularwidth}<{\centering} p{0.25\tabularwidth}<{\centering} p{0.25\tabularwidth}<{\centering} p{0.25\tabularwidth}<{\centering}}                        
\textit{Covariate} & \textit{Label} & \textit{Full} & \textit{No shift}\\  
& \expone\hspace{0.8mm}\exptwo\hspace{0.8mm}\expthree\hspace{0.8mm}		%
& 		%
& 		%

\vspace{2mm} \\
\end{tabular}\\
             
\rowcolor{gray!60}             
\textbf{Shift source} \\             
\footnotesize
\begin{tabular}{p{0.25\tabularwidth}<{\centering} p{0.25\tabularwidth}<{\centering} p{0.25\tabularwidth}<{\centering} p{0.25\tabularwidth}<{\centering}}                          
\textit{Naturally occurring} & \textit{Partitioned natural} & \textit{Generated shift} & \textit{Fully generated}\\
& 		%
& 		%
& \hspace{1.5em}\expone\hspace{0.8mm}\exptwo\hspace{0.8mm}\expthree\hspace{0.8mm}		%

\vspace{2mm} \\
\end{tabular}\\
             
\rowcolor{gray!60}             
\textbf{Shift locus}\\             
\footnotesize
\begin{tabular}{p{0.25\tabularwidth}<{\centering} p{0.25\tabularwidth}<{\centering} p{0.25\tabularwidth}<{\centering} p{0.25\tabularwidth}<{\centering}}                         
\textit{Train--test} & \textit{Finetune train--test} & \textit{Pretrain--train} & \textit{Pretrain--test}\\
& 		%
& 		%
& \hspace{1.5em}\expone\hspace{0.8mm}\exptwo\hspace{0.8mm}\expthree\hspace{0.8mm}		%

\vspace{2mm} \\
\end{tabular}\\

\hline
\end{tabular}
\caption{\textbf{Generator}'s evaluation card for the three main setups: baseline~\expone, self-aware adaptation~\exptwo, and audience-aware adaptation~\expthree }
\label{tab:eval-card-generator}
\end{table}
\renewcommand{\arraystretch}{1.1}         
\setlength{\tabcolsep}{0mm}    

\begin{table}[]
\centering
\begin{tabular}{|p{\tabularwidth}<{\centering}|}         
\hline
               
\rowcolor{gray!60}               
\textbf{Motivation} \\               
\footnotesize
\begin{tabular}{p{0.25\tabularwidth}<{\centering} p{0.25\tabularwidth}<{\centering} p{0.25\tabularwidth}<{\centering} p{0.25\tabularwidth}<{\centering}}                        
\textit{Practical} & \textit{Cognitive} & \textit{Intrinsic} & \textit{Fairness}\\
\exptwo\hspace{0.8mm}\expthree\hspace{0.8mm}		%
& \exptwo\hspace{0.8mm}\expthree\hspace{0.8mm}		%
& 		%
& 		%

\vspace{2mm} \\
\end{tabular}\\
               
\rowcolor{gray!60}               
\textbf{Generalisation type} \\               
\footnotesize
\begin{tabular}{m{0.17\tabularwidth}<{\centering} m{0.20\tabularwidth}<{\centering} m{0.14\tabularwidth}<{\centering} m{0.17\tabularwidth}<{\centering} m{0.18\tabularwidth}<{\centering} m{0.14\tabularwidth}<{\centering}}                   
\textit{Compo- sitional} & \textit{Structural} & \textit{Cross Task} & \textit{Cross Language} & \textit{Cross Domain} & \textit{Robust- ness}\\
& 		%
& 		%
& 		%
& \exptwo\hspace{0.8mm}\expthree\hspace{0.8mm}		%
& 		%

\vspace{2mm} \\
\end{tabular}\\
             
\rowcolor{gray!60}             
\textbf{Shift type} \\             
\footnotesize
\begin{tabular}{p{0.25\tabularwidth}<{\centering} p{0.25\tabularwidth}<{\centering} p{0.25\tabularwidth}<{\centering} p{0.25\tabularwidth}<{\centering}}                        
\textit{Covariate} & \textit{Label} & \textit{Full} & \textit{No shift}\\  
& \exptwo\hspace{0.8mm}		%
& 		%
& \hspace{2.5em}\expthree\hspace{0.8mm}		%

\vspace{2mm} \\
\end{tabular}\\
             
\rowcolor{gray!60}             
\textbf{Shift source} \\             
\footnotesize
\begin{tabular}{p{0.25\tabularwidth}<{\centering} p{0.25\tabularwidth}<{\centering} p{0.25\tabularwidth}<{\centering} p{0.25\tabularwidth}<{\centering}}                          
\textit{Naturally occurring} & \textit{Partitioned natural} & \textit{Generated shift} & \textit{Fully generated}\\
& 		%
& 		%
& \hspace{2em}\exptwo\hspace{0.8mm}\expthree\hspace{0.8mm}		%

\vspace{2mm} \\
\end{tabular}\\
             
\rowcolor{gray!60}             
\textbf{Shift locus}\\             
\footnotesize
\begin{tabular}{p{0.25\tabularwidth}<{\centering} p{0.25\tabularwidth}<{\centering} p{0.25\tabularwidth}<{\centering} p{0.25\tabularwidth}<{\centering}}                         
\textit{Train--test} & \textit{Finetune train--test} & \textit{Pretrain--train} & \textit{Pretrain--test}\\
& 		%
& 		%
& \hspace{2em}\exptwo\hspace{0.8mm}\expthree\hspace{0.8mm}		%

\vspace{2mm} \\
\end{tabular}\\

\hline
\end{tabular}
\caption{\textbf{Simulator}'s evaluation card for the two setups in which it is used (i.e., baseline setup excluded): self-aware adaptation~\exptwo, and audience-aware adaptation~\expthree.}
\label{tab:eval-card-simulator}
\end{table}
\renewcommand{\arraystretch}{1.1}         
\setlength{\tabcolsep}{0mm}    

\begin{table}[]
\centering
\begin{tabular}{|p{\tabularwidth}<{\centering}|}         
\hline
               
\rowcolor{gray!60}               
\textbf{Motivation} \\               
\footnotesize
\begin{tabular}{p{0.25\tabularwidth}<{\centering} p{0.25\tabularwidth}<{\centering} p{0.25\tabularwidth}<{\centering} p{0.25\tabularwidth}<{\centering}}                        
\textit{Practical} & \textit{Cognitive} & \textit{Intrinsic} & \textit{Fairness}\\
\expone\hspace{0.8mm}\exptwo\hspace{0.8mm}\expthree\hspace{0.8mm}		%
& \expone\hspace{0.8mm}\exptwo\hspace{0.8mm}\expthree\hspace{0.8mm}		%
& 		%
& 		%

\vspace{2mm} \\
\end{tabular}\\
               
\rowcolor{gray!60}               
\textbf{Generalisation type} \\               
\footnotesize
\begin{tabular}{m{0.17\tabularwidth}<{\centering} m{0.20\tabularwidth}<{\centering} m{0.14\tabularwidth}<{\centering} m{0.17\tabularwidth}<{\centering} m{0.18\tabularwidth}<{\centering} m{0.14\tabularwidth}<{\centering}}                   
\textit{Compo- sitional} & \textit{Structural} & \textit{Cross Task} & \textit{Cross Language} & \textit{Cross Domain} & \textit{Robust- ness}\\
& 		%
& 		%
& 		%
& \expone\hspace{0.8mm}\exptwo\hspace{0.8mm}\expthree\hspace{0.8mm}		%
& 		%

\vspace{2mm} \\
\end{tabular}\\
             
\rowcolor{gray!60}             
\textbf{Shift type} \\             
\footnotesize
\begin{tabular}{p{0.25\tabularwidth}<{\centering} p{0.25\tabularwidth}<{\centering} p{0.25\tabularwidth}<{\centering} p{0.25\tabularwidth}<{\centering}}                        
\textit{Covariate} & \textit{Label} & \textit{Full} & \textit{No shift}\\  
\expone\hspace{0.8mm}\exptwo\hspace{0.8mm}\expthree\hspace{0.8mm}		%
& 		%
& 		%
& \hspace{1.5em}\expone\hspace{0.8mm}\exptwo\hspace{0.8mm}\expthree\hspace{0.8mm}		%

\vspace{2mm} \\
\end{tabular}\\
             
\rowcolor{gray!60}             
\textbf{Shift source} \\             
\footnotesize
\begin{tabular}{p{0.25\tabularwidth}<{\centering} p{0.25\tabularwidth}<{\centering} p{0.25\tabularwidth}<{\centering} p{0.25\tabularwidth}<{\centering}}                          
\textit{Naturally occurring} & \textit{Partitioned natural} & \textit{Generated shift} & \textit{Fully generated}\\
& \expone\hspace{0.8mm}\exptwo\hspace{0.8mm}\expthree\hspace{0.8mm}		%
& 		%
& \hspace{1.5em}\expone\hspace{0.8mm}\exptwo\hspace{0.8mm}\expthree\hspace{0.8mm}		%

\vspace{2mm} \\
\end{tabular}\\
             
\rowcolor{gray!60}             
\textbf{Shift locus}\\             
\footnotesize
\begin{tabular}{p{0.25\tabularwidth}<{\centering} p{0.25\tabularwidth}<{\centering} p{0.25\tabularwidth}<{\centering} p{0.25\tabularwidth}<{\centering}}                         
\textit{Train--test} & \textit{Finetune train--test} & \textit{Pretrain--train} & \textit{Pretrain--test}\\
& 		%
& 		%
& \hspace{1.5em}\expone\hspace{0.8mm}\exptwo\hspace{0.8mm}\expthree\hspace{0.8mm}		%

\vspace{2mm} \\
\end{tabular}\\

\hline
\end{tabular}
\caption{\textbf{Listener}'s evaluation card for the three main setups: baseline~\expone, self-aware adaptation~\exptwo, and audience-aware adaptation~\expthree. In out-of-domain settings (\ood), the type of shift is covariate. In in-domain settings (\ind), there is no shift between the training and test.}
\label{tab:eval-card-listener}
\end{table}

\section{Additional Experiments}
\label{sec:app:exp}

Here, we provide details on additional experiments we performed in our adaptation pipeline.

In our adaptation mechanism, one of the stopping conditions is that the simulator predicts that the listener will be able to guess the referent. We also explored continuing adaptation until the \emph{listener} itself correctly guesses the referent. We report the results in Table \ref{tab:adapt_breaklist}, which reveal that using this stopping condition would yield higher results since the utterances are adapted until the actual listener makes a correct guess, mimicking an online interaction setup.

\begin{table}[]
\centering
\begin{tabular}{|l|ccc|}
\hline
\textbf{ Target domain  } & \textbf{ Golden  } & \textbf{ Speaker  } & \textbf{Adapted  } \\ \hline
\textbf{ appliances}                          & 16.85                    & 20.04                      & \textbf{38.89}                      \\ \hline
\textbf{ food}                                & 85.57                    & 55.26                      & \textbf{91.74}                      \\ \hline
\textbf{ indoor}                              & 18.69                    & 18.47                      & \textbf{39.49}                      \\ \hline
\textbf{ outdoor}                             & 19.03                    & 18.33                      & \textbf{37.96}                      \\ \hline
\textbf{ vehicles}                            & 13.75                    & 16.63                      & \textbf{35.43}                      \\ \hline
\end{tabular}
\caption{Listener accuracy using the listener stopping condition in the adaptation mechanism.}
\label{tab:adapt_breaklist}
\end{table}

\section{Additional Analyses}
\label{sec:app:analyses}

We note that we measure the type-utterance ratio for each step (i.e., the vocabulary size divided by the number of utterances available for that step), rather than the vocabulary size, because different steps correspond to different numbers of utterances: adaptation stops when the simulator module predicts the target image.%

\begin{figure}[h]
    \centering
    \includegraphics[width=0.8\columnwidth]{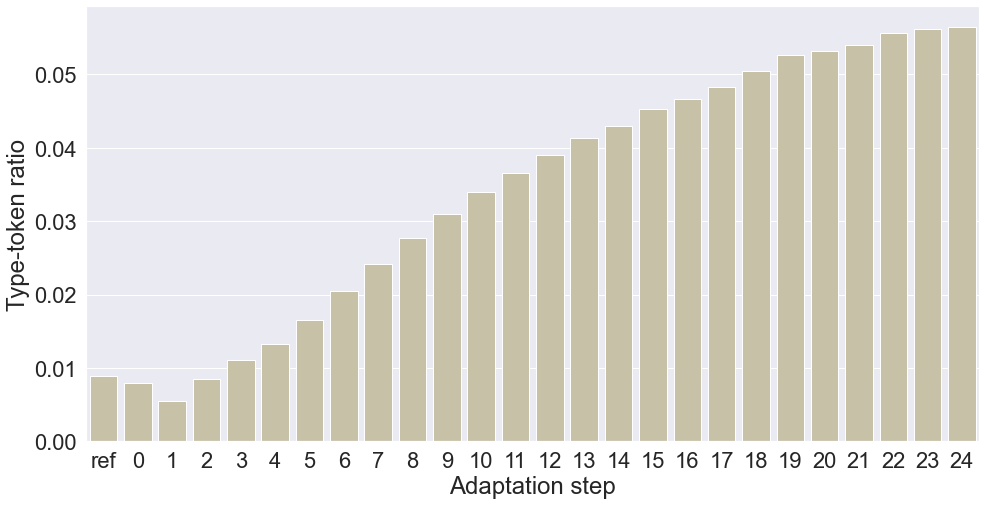}
    \caption{Type-token ratio across adaptation steps. Human gold utterances (\textit{ref}) and non-adapted utterances (0) also shown.}
    \label{fig:vocab-ttr}
\end{figure}

\begin{figure*}[ht]
    \centering
    \includegraphics[width=0.9\linewidth]{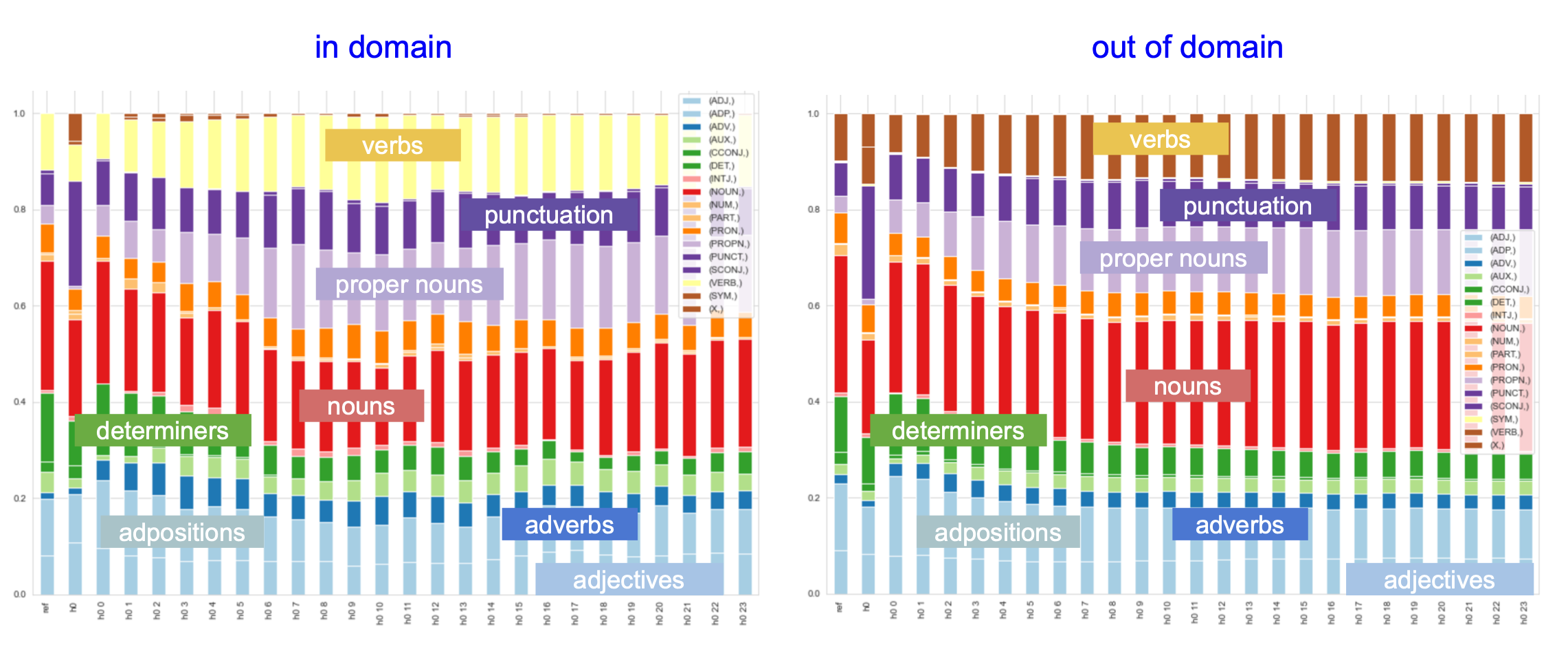}
    \caption{Unigram POS distribution across adaptation steps.}
    \label{fig:pos-ind-ood}
\end{figure*}

Figure~\ref{fig:pos-ind-ood} shows unigram part-of-speech distribution across adaptation steps for the in-domain and out-of-domain conditions.

\begin{figure}
    \centering
    \includegraphics[width=\linewidth]{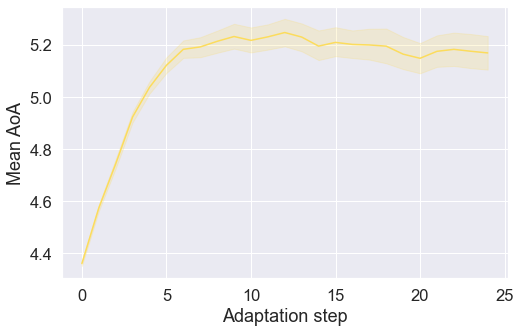}
    \caption{Mean utterance Age of Acquisition over adaptation steps. Step $0$ corresponds to the non-adapted utterance.}
    \label{fig:aoa-over-steps}
\end{figure}

\begin{figure}
    \centering
    \includegraphics[width=\linewidth]{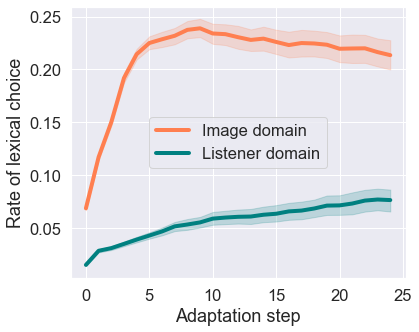}
    \caption{Rate of lexical choice from image and listener domain-specific vocabularies.}
    \label{fig:domain-over-steps}
\end{figure}

We also measure the domain-specificity of utterances over steps, both in terms of the target image domain and of the listener domain, as the percentage of domain-specific words in an utterance. 
We consider as domain-specific the words that appear \textit{only} in interactions about a certain domain. 
The speaker, throughout adaptation, produces more words belonging to both the image and the listener domain (Figure~\ref{fig:domain-over-steps}) and thus less domain-agnostic words.
We saw that, over adaptation steps, the decoder hidden state forgets image domain information in favour of the listener domain. This does not translate into no longer producing words from the image domain, suggesting that the speaker may be focusing more on the specific image than on its semantic domain. %

Figure~\ref{fig:aoa-over-steps} shows mean utterance age of acquisition rating \cite{kuperman2012age} over steps.

\end{document}